\documentclass[lettersize,journal]{IEEEtran}
\usepackage{amsmath,amsfonts}
\usepackage{algorithmic}
\usepackage{algorithm}
\usepackage{array}
\usepackage[caption=false,font=normalsize,labelfont=sf,textfont=sf]{subfig}
\usepackage{textcomp}
\usepackage{stfloats}
\usepackage{url}
\usepackage{verbatim}
\usepackage{graphicx}
\usepackage{cite}
\usepackage{setspace}
\usepackage{multirow}
\usepackage{makecell}
\usepackage{booktabs}
\usepackage{color}
\usepackage{url}
\usepackage{siunitx}
\usepackage{booktabs}
\usepackage{amsmath} 
\usepackage{arydshln}

\begin{document}

\title{Mitigating Low-Level Visual Hallucinations Requires\\ Self-Awareness: Database, Model and Training Strategy}

\author{Yinan~Sun,
        Xiongkuo~Min$^{*}$,~\IEEEmembership{Member,~IEEE,}
        Zicheng~Zhang, \\
        Yixuan~Gao,
        Yuqin~Cao,        and~Guangtao~Zhai$^{*}$,~\IEEEmembership{Fellow,~IEEE}
\thanks{*Corresponding authors: Xiongkuo~Min and Guangtao~Zhai.}
}

\markboth{Journal of \LaTeX\ Class Files,~Vol.~14, No.~8, August~2021}
{Shell \MakeLowercase{\textit{et al.}}: A Sample Article Using IEEEtran.cls for IEEE Journals}

\maketitle

\begin{abstract}
The rapid development of multimodal large language models has resulted in remarkable advancements in visual perception and understanding, consolidating several tasks into a single visual question-answering framework. However, these models are prone to hallucinations, which limit their reliability as artificial intelligence systems. While this issue is extensively researched in natural language processing and image captioning, there remains a lack of investigation of hallucinations in Low-level Visual Perception and Understanding (HLPU), especially in the context of image quality assessment tasks. We consider that these hallucinations arise from an absence of clear self-awareness within the models. To address this issue, we first introduce the HLPU instruction database, the first instruction database specifically focused on hallucinations in low-level vision tasks. This database contains approximately 200K question-answer pairs and comprises four subsets, each covering different types of instructions. Subsequently, we propose the Self-Awareness Failure Elimination (SAFEQA) model, which utilizes image features, salient region features and quality features to improve the perception and comprehension abilities of the model in low-level vision tasks. Furthermore, we propose the Enhancing Self-Awareness Preference Optimization (ESA-PO) framework to increase the model's awareness of knowledge boundaries, thereby mitigating the incidence of hallucination. Finally, we conduct comprehensive experiments on low-level vision tasks, with the results demonstrating that our proposed method significantly enhances self-awareness of the model in these tasks and reduces hallucinations. Notably, our proposed method improves both accuracy and self-awareness of the proposed model and outperforms close-source models in terms of various evaluation metrics. This research contributes to the advancement of self-awareness capabilities in multimodal large language models, particularly for low-level visual perception and understanding tasks.
\end{abstract}

\begin{IEEEkeywords}
Multimodal large language models, low-level vision, image quality assessment, hallucination.
\end{IEEEkeywords}

\section{Introduction}
In recent years, the rapid development of Large Language Models (LLMs), including close-source models such as GPT and Bard, as well as open-source models like LLaMA and MPT, has significantly influenced the trajectory of artificial intelligence (AI), ushering in a new era of AI-assisted living. To better align with diverse usage scenarios, researchers have integrated pretrained vision encoders with LLMs and subsequently finetuned these combinations, leading to the emergence of Multimodal Large Language Models (MLLMs). Benefiting from large-scale pre-training databases \cite{laurenccon2024obelics} and high-quality supervised finetuning data \cite{chen2024far, dai2023instructblip}. MLLMs exemplified by models such as LLaVA and InstructBLIP have demonstrated their considerable potential in various domains \cite{alayrac2022flamingo, bampis2017study}, including image captioning, visual question answering, image segmentation, medical image diagnosis, image quality assessment (IQA) and aesthetic evaluation \cite{gao2024no, ding2020image, min2019quality, hosu2020koniq, gao2023blind, min2025exploring, cao2023attention, cao2024unqa, cao2025agav, gao2022image}.

\begin{figure}[!t]
\centerline{\includegraphics[width=3in]{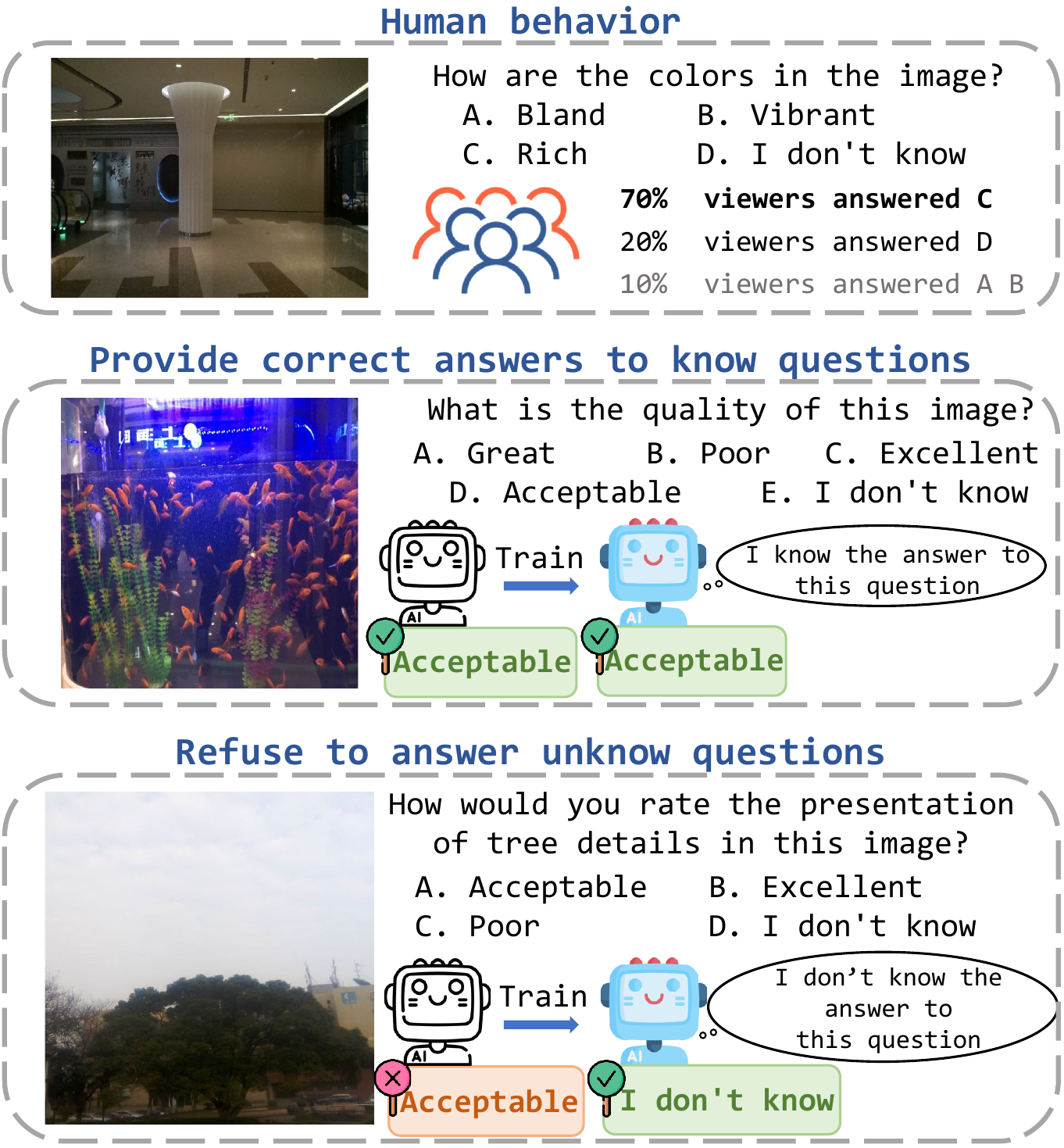}}
\vspace{-1em}
\caption{The self-awareness of models in low-level vision tasks. A reliable model should be able to accurately recognize what it knows and what it does not know. It should provide correct answers to questions within its basic knowledge, while declining to answer questions that fall outside of its basic knowledge.}
\label{example}
\vspace{-1.5em}
\end{figure}

Despite the rapid advancements in MLLMs, they are still not regarded as fully reliable AI systems. A major challenge impeding their practical application is hallucination \cite{wei2023privileged, lin2021joint}. Hallucination refers to the inconsistency between the actual content of the image and the text generated by the model, resulting in answers that may be factually incorrect or nonsensical \cite{fang2024linguistic}. This phenomenon significantly undermines the reliability and applicability of MLLMs. The occurrence of hallucinations has been a persistent issue in LLMs. One contributing factor to this phenomenon is the limited self-awareness of these models. While LLMs are trained on diverse databases to acquire extensive knowledge, these databases fail to sensitize models to the boundaries of their knowledge, meaning that models are not equipped with ability to distinguish between known factual domains and uncharted information territories. Consequently, LLMs often lack a clear understanding of what they know and what they do not know. When confronted with unfamiliar information, LLMs tend to generate responses that are either nonsensical or outdated \cite{huang2024survey}. To mitigate hallucinations in LLMs, researchers have proposed various methods, including inference-stage strategies like contrastive decoding \cite{leng2024mitigating}, as well as the utilization of external visual models for post-hoc response correction \cite{yin2024woodpecker}. While these methods are simple and training-free, they do not fundamentally enhance the intrinsic capabilities of the models. More recently, researchers have introduced some pioneering preference optimization methods, such as Direct Preference Optimization (DPO) \cite{rafailov2024direct}, which encourage models to learn from comparisons between positive and negative samples, offering a promising avenue for reducing hallucinations \cite{pi2024strengthening}.

Preliminary research on hallucinations in MLLMs has primarily focused on high-level vision tasks. Researchers have utilized various methods, including Chain of Thought (COT), Direct Preference Optimization (DPO), Proximal Policy Optimization (PPO) and Reinforcement Learning from Human Feedback (RLHF) to mitigate hallucination effects, particularly in visual question answering tasks, yielding promising results. However, the challenges associated with hallucinations in the low-level vision tasks domain remain largely unexplored. Meanwhile, the self-awareness of MLLMs in low-level visual perception and understanding plays significant roles in IQA \cite{hosu2020koniq, fang2020perceptual} and related tasks, which involve perceptual distortions (such as noise and blur) \cite{su2021koniq++, wu2023towards} and other low-level attributes (\textit{e.g.}, color, lighting, composition and style) \cite{kong2016photo} related to the aesthetics of natural photographs \cite{murray2012ava}, computer graphics images \cite{zhang2023subjective} and AI generated images \cite{xu2024imagereward, li2023agiqa}. These low-level visual capabilities are important for a range of applications, including recommendation systems \cite{wu2023exploring}, camera guidance \cite{zhang2022exploring} and visual quality enhancement \cite{zhang2018unreasonable}. Consequently, it is important to evaluate the self-awareness capabilities of MLLMs in low-level visual perception and understanding, to indirectly assess the credibility of their responses to specific low-level tasks.

One potential cause of hallucinations is the lack of knowledge or context relevant to the given task of models. For tasks involving fundamental visual perception and understanding, a promising solution for MLLMs is to respond with ``I don't know'' when confronted with a question that exceeds their basic knowledge or the context provided. However, researchers have noted that LLMs often struggle to recognize their limitations in knowledge \cite{fang2020perceptual}, a phenomenon that is similarly observed in MLLMs. As illustrated in Fig. \ref{example}, we hope that MLLMs behave more like humans. The self-awareness of these models should empower them to provide accurate responses with confidence and in scenarios where questions exceed the available visual information or the model's knowledge scope, to refrain from providing an answer rather than delivering wrong or misleading answers.

In this work, we aim to conduct an in-depth analysis of the self-awareness capabilities of MLLMs in low-level visual perception and understanding, build an accurate model for low-level tasks and propose a training framework, which can enhance the self-awareness capabilities of models. To achieve this objective, we are facing the following research challenges.

\textbf{(i)} Building an instruction database for low-level visual perception and understanding in order to mitigate hallucinations. Although there are many publicly available low-level visual databases, such as KonIQ-10k \cite{hosu2020koniq}, AGIQA-3k \cite{li2023agiqa}, Q-Instruct \cite{wu2024q}, \textit{etc.}, they don't enhance the self-awareness capabilities of models. In addition, some other databases, such as Visual Genome (VG) \cite{krishna2017visual}, which is used to mitigate the occurrence of hallucinations, focus on high-level vision tasks rather than low-level vision tasks.

\textbf{(ii)} Understanding the phenomenon of hallucinations in different low-level visual contexts. As a common observation, the phenomena of hallucinations have many manifestations, including the fabrication of facts and the inconsistencies between texts and images. Different types of tasks exhibit varying degrees of hallucination. However, whether different low-level visual features influence these phenomena of hallucinations still remains unknown.

\textbf{(iii)} Constructing a model for low-level visual perception and understanding tasks and proposing a framework to enhance the self-awareness capabilities of the model. As MLLMs increasingly become an indispensable part of daily life, their reliability has become a focal point of concern. It is necessary and significant to propose a model that focuses on low-level visual features and a universal training framework, which aims to improve the accuracy and reliability of the responses generated by MLLMs in low-level vision tasks.

In this work, we first construct a database focusing on the hallucination in low-level visual perception and understanding tasks, termed the HLPU instruction database. Specifically, as shown in Fig. \ref{pipeline}, our database construction pipeline is segmented into three phases, comprising approximately 200K samples, each including an image accompanied by three types of responses. To better explore the self-awareness capabilities of MLLMs, we propose a novel model (SAFEQA) and a universal train framework (ESA-PO), which can be used to improve the accuracy and enhance the self-awareness abilities of models. Specifically, we utilize a salient region feature extraction module and a quality feature extraction module to enhance the ability of model to extract low-level visual features. Additionally, we integrate preference settings, using ``I don't know'' as a suboptimal preference into the model. We construct the LLSAVisionQA database as the test set. Experimental results demonstrate the effectiveness of the proposed method in improving the answer accuracy and self-awareness ability of MLLMs in low-level visual perception and understanding tasks. The contributions are summarized as follows:

\begin{itemize}
\item We build a new low-level visual perception and understanding instruction database focusing on hallucination, termed as the HLPU instruction database, which aims to study the self-awareness capabilities of models in low-level vision tasks. 
\item We analyze the self-awareness capabilities of models across different types of questions and various low-level visual features and indicate that self-awareness capabilities are significantly influenced by question and distortion type. 
\item We validate the performance of state-of-the-art MLLMs on our LLSAVisionQA database and establish a benchmark regarding low-level visual perception and understanding.
\item A low-level visual perception and understanding model and a training framework are proposed. The model demonstrates a better perception of low-level visual features compared to the benchmark methods. Experimental results show that the training framework enhances the self-awareness capabilities of models in low-level visual perception and understanding, reducing the occurrence of hallucinations and outperforming the benchmark methods.
\end{itemize}

The rest of the paper is organized as follows. Section \ref{Related Work} introduces related works. Then we introduce the construction procedure of the HLPU instruction database in Section \ref{Database}. Section \ref{Proposed Methods} introduces our proposed SAFEQA model and ESA-PO framework in detail. Section \ref{Experiments} describes the experimental settings and the experimental results. Section \ref{Conclusion} concludes the whole paper.

\section{Related Work}
\label{Related Work}
\subsection{Low-level Visual Perception and Understanding}
Image quality assessment has long been recognized as a fundamental task in the comprehension and interpretation of visual content, with the objective of predicting quality scores that accurately reflect human visual perception. Over the past two decades, various IQA databases have been established to support the development of IQA algorithms for both general images and specific domains, including artificially-distorted images \cite{lin2019kadid} (\textit{e.g.}, JPEG, AWGN, \textit{etc.}), in-the-wild images \cite{hosu2020koniq, ying2020patches} and recently-popular AI-generated images \cite{xu2024imagereward, li2023agiqa}. These databases provide important metrics for the production and distribution of visual content. Despite general IQA, recent studies have started to focus on more finer-grained low-level visual aspects and explored some related tasks such as evaluating low-level visual dimensions (\textit{e.g.}, color and brightness) \cite{fang2020perceptual, wu2023towards}, or distinguishing distortions (\textit{e.g.}, blurriness, noise and overexposure) \cite{su2021koniq++}. Recent studies \cite{wu2023exploring2, wu2023towards, wu2023exploring} have even considered image-related dimensions, such as composition, lighting and bokeh, as low-level aspects in a broader sense \cite{kong2016photo}. In general, low-level visual perception and understanding tasks encompass all aspects of image appearance (in contrast to object-level contents) that can be perceived by human and evoke various sensory responses.

\subsection{Low-level Visual Database}
While low-level vision tasks were previously solved separately, the recently proposed databases have covered most low-level visual information, providing an opportunity to integrate these capabilities into a foundational model. Q-Bench \cite{zhang2024benchmark} is an image-centered visual question-answering benchmark designed to evaluate the low-level visual perception and understanding tasks of MLLMs. Q-Instruct \cite{wu2024q} has constructed a comprehensive instruction database through human annotations, significantly enhancing the ability of MLLMs to understand the low-level visual of images. Co-Instruct \cite{wu2024towards} has created instruction databases through multi-image comparison and joint analysis, significantly improving large models' ability to analyze the quality of multiple visual stimuli. DepictQA \cite{you2024depicting} has employed MLLMs to provide detailed, language-driven evaluations, outperforming traditional score-based methods. AesExpert \cite{huang2024aesexpert} has proposed expert-level aesthetic foundational models by assembling an extensive database of image aesthetic reviews. However, despite these advances, the phenomena of hallucination still persist and there has been a lack of research investigating the self-awareness capabilities of MLLMs. Our work is the first to investigate the self-awareness capabilities of MLLMs in low-level visual perception and understanding tasks, thus establishing a promising way for mitigating hallucination through MLLMs.

\subsection{Mitigating Hallucinations in MLLMs}
VCD \cite{leng2024mitigating} mitigates the issue of model hallucinations by reducing knowledge bias through comparative decoding. While this method is notably effective, it requires the model to perform two inferences for each token prediction, resulting in increased memory consumption and higher management costs in real-world applications. Opera \cite{huang2024opera} mitigates hallucinations by identifying common patterns in decoding attention scores when modeling hallucinations and applying special decoding strategies. However, this method increases the inference load and decelerates processing speed. HA-DPO \cite{zhao2023beyond} views hallucinations as the preference of the model. By leveraging ChatGPT \cite{achiam2023gpt} and ground truth annotations from existing image databases, it generates positive instructions that align with image content, using the original output of the model as negative instructions for direct preference optimization. Although this method is effective, the construction of negative instructions may not be optimal, as it might not comprehensively encompass the diverse forms of hallucinations. Similarly, BPO \cite{pi2024strengthening} not only adds noise but also injects errors into positive instructions through the LLM backbone of MLLMs to construct negative instructions. These methods are limited to positive and negative pairs and our paper extends them to enhance both accuracy and self-awareness of the model.

\begin{figure*}[!t]
\centering
\includegraphics[width=6.5in]{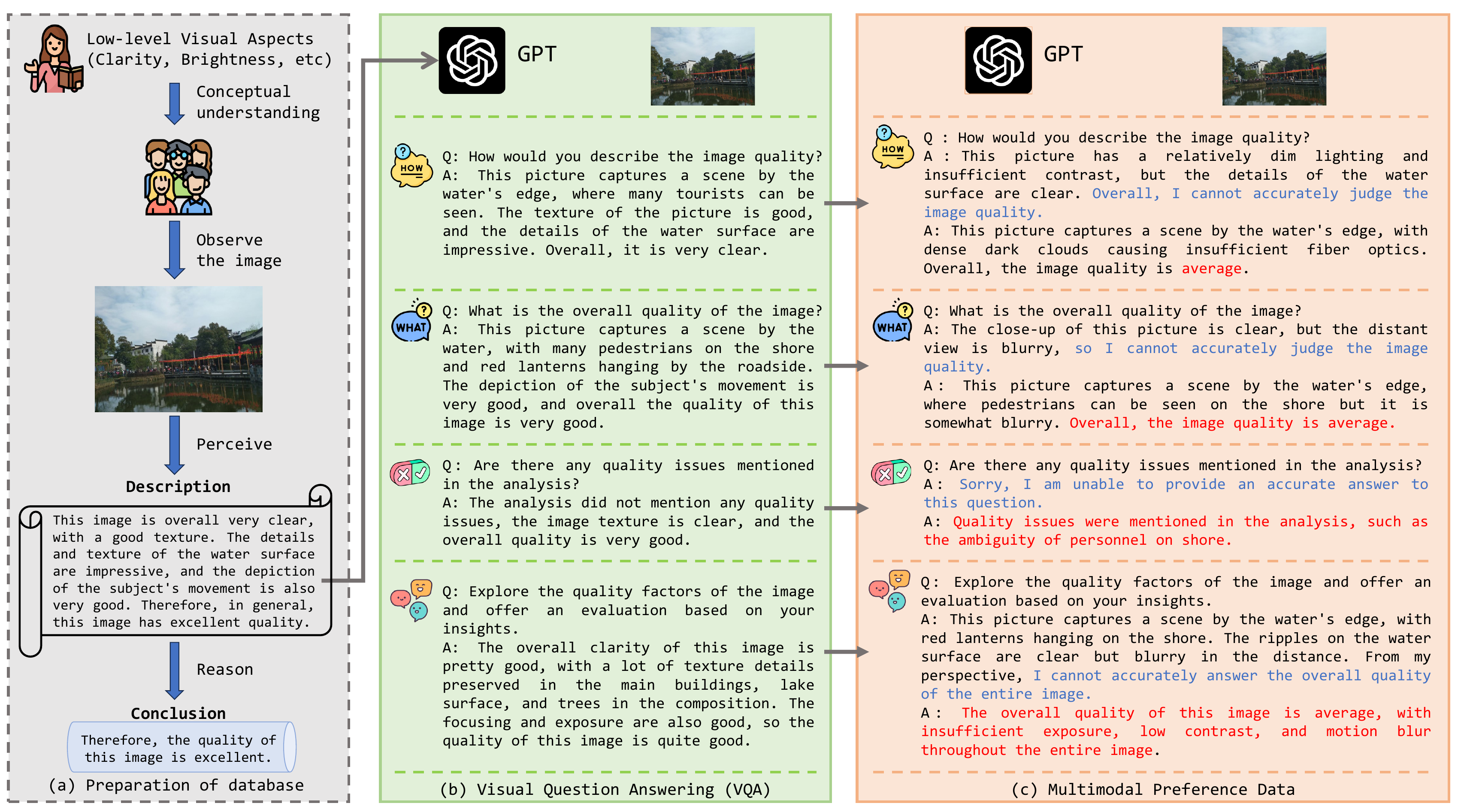}
\vspace{-1em}
\caption{HLPU instruction database construction pipeline. First, we select 76K available data as the original data and the subjective experimental process is shown in (a) Preparation of database. Secondly, we use GPT to convert the database into 200K instruction-response pairs and provide as detailed answers as possible for each question, as shown in (b) Visual question answering (VQA). Finally, we use GPT to generate multimodal preference data, which are used for low-level visual preference optimization training strategy, as shown in (c) Multimodal preference data.}
\label{pipeline}
\vspace{-1em}
\end{figure*}

\section{Database}
\label{Database}
Due to the absence of a comprehensive database for low-level visual hallucinations, we construct the first low-level visual perception and understanding instruction database focusing on hallucination, denoted as the HLPU instruction database. We build an extensible data construction pipeline, as illustrated in Fig. \ref{pipeline}, to obtain more diverse multimodal preference data. The database construction is composed of three stages: In \textit{Stage-1}, we select relevant data from the Q-Instruct database (Sec. \ref{Preparation of Database}). \textit{Stage-2} focuses on expanding the available data (Sec. \ref{VQA}), while \textit{Stage-3} is dedicated to generating multimodal preference data (Sec. \ref{Multimodal Preference Data}). Finally, we introduce the LLSAVisionQA database for evaluative purposes (Sec. \ref{test data}).

\subsection{Database Preparation}
\label{Preparation of Database}
To ensure the diversity of instructions and images, the original data for the HLPU instruction database is sourced from the Q-Pathway and the extended conversations of the Q-Instruct database, termed as the L-Base database. The images are sampled from various sources, including four in-the-wild IQA databases \cite{fang2020perceptual, ghadiyaram2015massive, hosu2020koniq, ying2020patches} and two databases with AI-generated images \cite{xu2024imagereward, li2023agiqa}. Specifically, the L-Base database is carefully constructed sub-sampled image groups from the original databases to introduce more diversity in low-level visual appearances, achieving a balance between high-quality and low-quality images. Additionally, in the L-Base database, researchers randomly introduced artificial distortions into 1,012 original images from the COCO database \cite{chen2015microsoft}, to simulate real-world quality issues. The instructions in the L-Base database consist of two parts. The first part includes detailed natural language descriptions of low-level visual attributes (\textit{e.g.}, noise, brightness and clarity) provided by human subjects, along with general conclusions. It is important to note that, as the majority of images come from IQA databases, the mean opinion scores (MOSs) of them are also displayed to the subjects to better calibrate them with a common understanding of quality, thereby ensuring the accuracy of the descriptions provided. The second part of the instructions consists of extended conversations based on the human-provided descriptions, aimed at improving the ability to discuss with human grounded on the low-level visual aspects regarding the input image. In the L-Base database, we refer to the image as $I$ and the instruction description as $Y_{p}$.

\begin{figure*}[!t]
\centering
\includegraphics[width=6.5in]{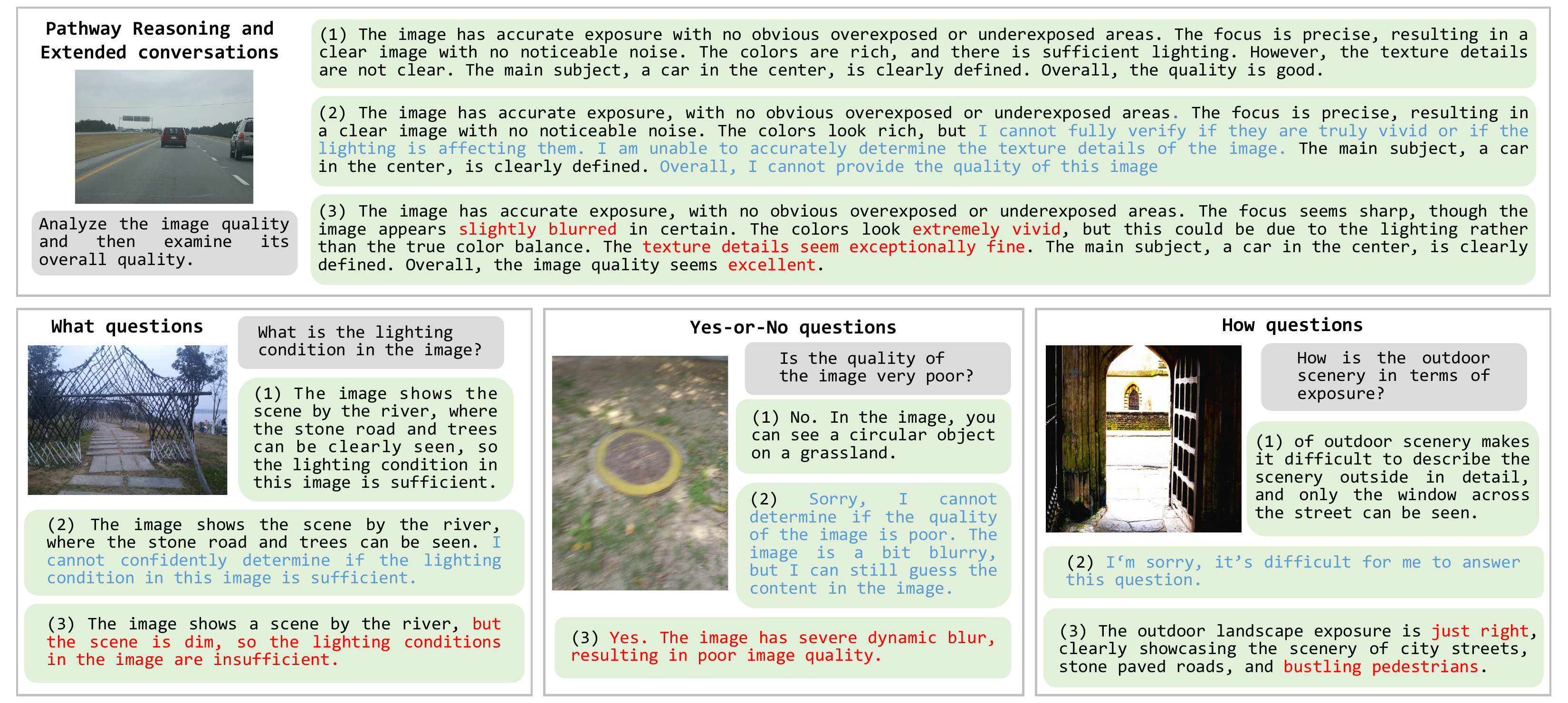}
\vspace{-1em}
\caption{The composition of the HLPU instruction database, in which the 200K instruction-response pairs include (a) Pathway reasoning and extended conversations, (b) ``What'' questions, (c) ``Yes-or-No'' questions and (d) ``How'' questions. The blue part of the responses indicates refusal and the red part of the responses indicates the hallucination.}
\label{database}
\vspace{-1em}
\end{figure*}

\subsection{Visual Question Answering (VQA)}
\label{VQA}
Before generating multimodal preference data, we design a pipeline utilizing GPT-4O \cite{achiam2023gpt} to convert ($i$, $y_{p}$) pairs in the L-Base database into visual question answering (VQA) subsets. Specifically, we require GPT-4O to generate diverse-style questions related to low-level vision based on L-Base and provide detailed answers whenever possible. This process results in 76K questions, which include ``how'' questions answered with opinion-related adjectives (\textit{e.g.} good/poor and high/low), or ``what'' questions answered with attribute-related (blur/noise/focus) or context-related (left/the peacock/the background) nouns, as illustrated in Fig. \ref{database}. Additionally, we instruct GPT to generate binary judgment questions, as shown in Fig. \ref{database}, ensuring that the yes/no responses are balanced at 1:1 ratio. This results in a collection of 57K yes-or-no questions. The generated questions are recorded as X, with the corresponding answers as $Y_{p}$. Using this pipeline, \textit{Stage-2} results in the construction of a large-scale multimodal database for low-level visual features, containing approximately 200K ($i$, $x$, $y_{p}$) sample pairs, each focused on different aspects of low-level vision tasks.

\subsection{Multimodal Preference Data}
\label{Multimodal Preference Data}
To construct multimodal preference data, we propose a straightforward and effective method known as the mask token prediction method. Each sample in the HLPU instruction database consists of an image \( i \in I \), an instruction \( x \in X \), a positive response as \( y_{p} \in Y_{p} \), a suboptimal response \( y_{d} \in Y_{d} \) and a negative response \( y_{n} \in Y_{n} \). To enhance the self-awareness ability of models, it is essential to construct a suboptimal response set $Y_{d}$ and a negative response set $Y_{n}$, where $Y_{p}$ is preferable to $Y_{d}$ and $Y_{d}$ is preferable to $Y_{n}$. This relationship is transitive. Given a certain image $i$, instruction $x$ and positive response $y_{p}$, we derive the suboptimal response and negative response from an initial instruction model as follows:
\begin{equation}
y_d, y_n \sim M(y_p \mid x, i),
\end{equation}
where $M$ represents the model, $i$ represents the image, $y_{p}$, $y_{d}$ and $y_{n}$ represent the positive response, suboptimal response and negative response, respectively. We categorize human responses, extended conversations and GPT-generated responses as elements of the positive response set $Y_{p}$. To generate a suboptimal response set $Y_{d}$ and a negative response set $Y_{n}$, we obtain responses from $Y_{p}$ and mask certain positions. The model $M$ is required to complete the masked portion of the response without using the image to derive the negative response set $Y_{n}$. Alternatively, we rephrase the masked segments as ``I don't know'', to obtain the suboptimal response set $Y_{d}$. Using this pipeline, in \textit{Stage-3}, we construct multimodal preference data. As illustrated in Fig. \ref{database}, each ($i$, $x$, $y_{p}$) sample pair is supplemented with a generated suboptimal response $y_{d}$ and a negative response $y_{n}$. Ultimately, the HLPU instruction database contains approximately 200K ($i$, $x$, $y_{p}$, $y_{d}$, $y_{n}$) sample pairs, which focus on low-level visual perception and understanding.

\subsection{LLSAVisionQA Database}
\label{test data}
To evaluate the self-awareness capabilities of different multimodal large language models on various low-level attributes under diverse circumstances, we construct the LLSAVisionQA database, which consists of 2,990 images sourced from 10 diverse sources. Aligned with existing research \cite{liu2023mmbench, lu2023evaluation}, each image in the LLSAVisionQA database is accompanied by a question, a correct answer, false candidate answers and an ``I don't know'' option. Specifically, we have designed three diverse types of questions: Yes-or-No questions, What questions and How questions. Furthermore, we divide low-level concerns for images into four quadrants via two axes: (1) distortions (blur, noises, \textit{etc.}) and other low-level attributes (color, lighting, composition, \textit{etc.}) \cite{guha2020atqam}. (2) global perception (\textit{e.g.}, sharpness of the whole picture) and local in-context perception (\textit{e.g.}, whether the red flower is in focus) \cite{li2018has}. The LLSAVisionQA database provides a comprehensive benchmark for evaluating the self-awareness abilities of MLLMs in low-level visual perception and understanding tasks, covering aspects such as distortion, color and overall visual attributes through the use of these three question types and four low-level concerns.

\begin{figure*}[!t]
\centering
\includegraphics[width=6in]{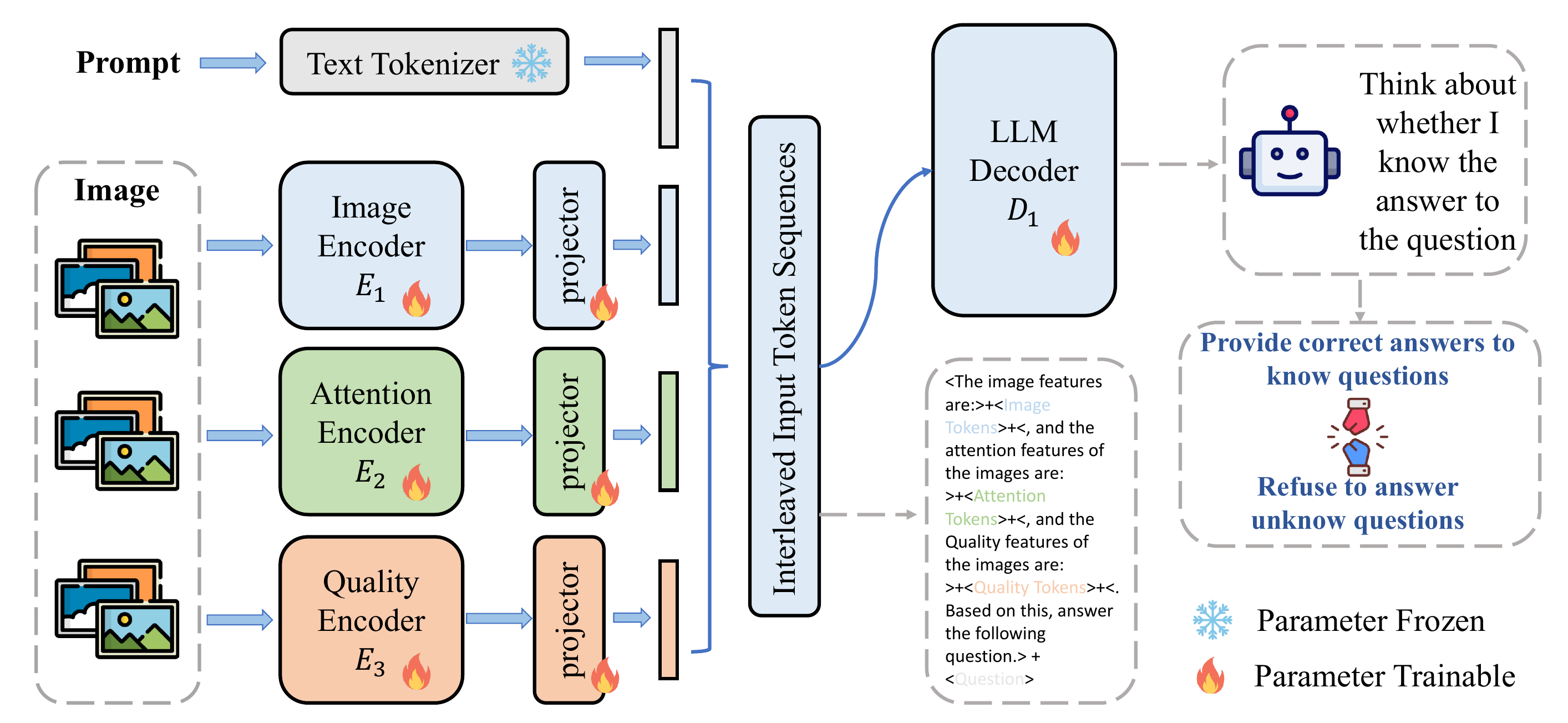}
\vspace{-1em}
\caption{The detailed architecture of the proposed model, which is composed of five modules. (a) The image feature extraction module extracts image features using SigLIP. (b) The salient region feature extraction module extracts salient features using Swin Transformer. (c) The quality feature extraction module extracts quality features using Swin Transformer. (d) The text tokenizer. (e) The LLM decoding module integrates features and outputs responses.}
\label{model}
\vspace{-1.5em}
\end{figure*}

\section{Proposed Method}
\label{Proposed Methods}
\begin{figure*}[!t]
\centering
\includegraphics[width=6in]{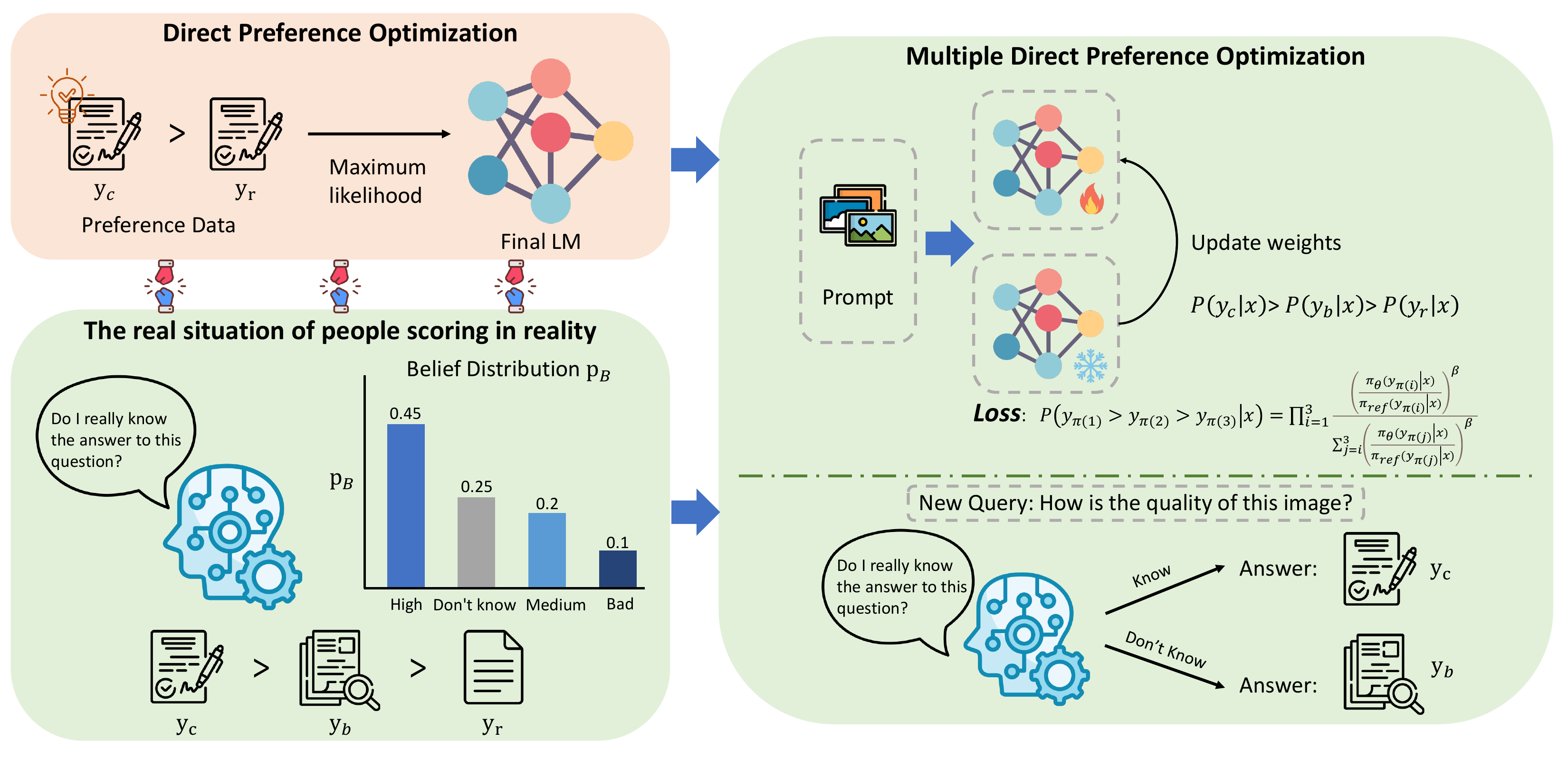}
\vspace{-1em}
\caption{The detailed structure of the proposed training strategy. The traditional DPO method includes positive responses and negative responses. But in real-life situations, the response to the unknown question is often ``I don't know''. Our proposed method incorporates ``I don't know'' as a suboptimal response to enhance the self-awareness ability of the models.}
\label{esa-po}
\vspace{-1em}
\end{figure*}

In this section, we describe the architecture of our proposed model, SAFEQA, as well as the training strategy designed to mitigate hallucinations, ESA-PO. The overall structure of the model is shown in Fig. \ref{model}, consisting of five main modules: a image feature extraction module, a salient region feature extraction module, a quality feature extraction module, a text tokenizer and a LLM decoding module, as described in Section \ref{Model Structure}. The training strategy is illustrated in Fig. \ref{esa-po}. We first introduce the foundational concepts of the training strategy in Section \ref{DPO method} and then offer a comprehensive explanation of our proposed training strategy ESA-PO in Section \ref{our method}.

\subsection{SAFEQA Model Structure}
\label{Model Structure}
\textbf{Base Model.} We select LLaVA-OneVision-Chat-7B \cite{li2024llava} as the foundation of our model. The foundation model achieves excellent performance on various high-level visual question answering benchmarks \cite{song2023recovering, guo2024unk} and demonstrates remarkable capabilities in image semantic understanding and reasoning. The foundational model comprises a visual tower constructed from SigLIP \cite{zhai2023sigmoid}, which is employed to extract feature tokens from images, a vision projector composed of fully connected layers for feature mapping and the Qwen-2 model \cite{yang2024qwen2}, whose tokenizer serves as the LLM and text embedding layers.

\textbf{Low-level Visual Feature Extraction Module.} While LLaVA-OneVision-Chat-7B \cite{li2024llava} has shown impressive results on high-level visual question answering tasks, we believe its performance in low-level visual question answering could be further improved. Given that one of the primary tasks in low-level visual perception and understanding is visual quality assessment, we incorporate additional modules to enhance the feature extraction capability of the model. Specifically, we use a salient region feature extraction module and a quality evaluation feature extraction module to improve the model's ability to process low-level visual information. We then employ a salient projector and a quality projector both identical in structure to the vision projector to map the tokens. This ensures consistency in dimensions between visual and text tokens. During training, the token sequences are input in an interleaved manner.

\subsection{Direct Preference Optimization Training}
\label{DPO method}
To better align LLMs with human preferences, various preference optimization methods are developed. Reinforcement Learning with Human Feedback (RLHF) \cite{christiano2017deep} has emerged as a prominent technique in aligning LLMs \cite{bai2022training, ouyang2022training, stiennon2020learning}. However, it involves the training of reward models, which presents significant challenges. In contrast, Rafailov \textit{et al.} \cite{rafailov2024direct} introduced the Direct Preference Optimization (DPO) method, which directly utilized preference data without requiring reward models and demonstrated impressive performance \cite{ivison2023camels, tunstall2023zephyr}. The naive DPO uses a pair of preference data and is based on the Bradley-Terry model for optimizing the large language model. The loss function for DPO is defined as follows:
\begin{align}
\mathcal{L}_d 
&= -\,\mathbb{E}_{\mathcal{D}} \Bigl[ 
  \log \sigma \Bigl( 
    \beta \,\log \frac{\pi_\theta(y_c \mid x)}{\pi_{\text{ref}}(y_c \mid x)}
    \;-\;
    \beta \,\log \frac{\pi_\theta(y_r \mid x)}{\pi_{\text{ref}}(y_r \mid x)}
  \Bigr) 
\Bigr],
\end{align}
where $x$ is the input prompt, $y_c$ is the positive response, $y_r$ is the negative response, $\pi_{\text{ref}}$ represents the reference policy, $\pi_\theta$ represents the policy being optimized, $D$ is the database of prompts and responses, $\sigma$ is the sigmoid function and $\beta$ is the scaling parameter. By directly incorporating preference data into the optimization process, DPO effectively ensures that the generated text aligns closely with human judgments.

\subsection{ESA-PO Training Strategy}
\label{our method}
In our proposed HLPU instruction database, each sample is associated with three distinct responses: $Y_p$, $Y_d$ and $Y_n$. $Y_p$ represents the positive response, $Y_d$ represents the suboptimal response and $Y_n$ represents the negative response. The primary objective of our work is to develop a model that emphasizes low-level visual perception and comprehension, simulate human responses and mitigate the occurrence of incorrect answers. To accomplish this aim, we propose the preference optimization method, ESA-PO, for enhancing self-awareness abilities, which utilizes preference data to better align multimodal models with human preferences. To simultaneously utilize the three types of preferences in the database, we employ the Plackett-Luce model \cite{plackett1975analysis} to characterize the ranking probabilities. The definition of ranking probability is derived from the product of sequentially selected conditional probabilities, which is formulated as:
\begin{equation}
p(y_1 \succ y_2 \succ y_3 \mid x) = p(y_1 \mid x) \cdot p(y_2 \mid y_1, x),
\end{equation}
where $x$ represents the input, $y_1$ represents the positive response, $y_2$ represents the suboptimal response and $y_3$ represents the negative response. The probability of selecting the positive answer, the conditional probability of the suboptimal answer and the complete ranking probability are formulated as:
\begin{equation}
p(y_1 \mid x) 
= \frac{\exp\bigl(r(x, y_1)\bigr)}{\sum_{j=1}^{3} \exp\bigl(r(x, y_j)\bigr)},
\end{equation}
\begin{equation}
p(y_2 \mid y_1, x) = \frac{\exp(r(x, y_2))}{\exp(r(x, y_2)) + \exp(r(x, y_3))},
\end{equation}
\begin{equation}
p(y_1 \succ y_2 \succ y_3 \mid x) = p(y_1 \mid x) \cdot p(y_2 \mid y_1, x),
\end{equation}
where $r(x,y)$ is the reward function, which can be represented by the policy distribution, reference policy and scaling factor. The reward function is formulated as:
\begin{equation}
r(x, y) = \beta \log \frac{\pi_\theta(y \mid x)}{\pi_{\text{ref}}(y \mid x)},
\end{equation}
where $\pi_{\text{ref}}$ represents the reference policy, $\pi_\theta$ represents the policy being optimized and $\beta$ represents the scaling parameter. By substituting the ranking probability formula, the final loss function is formulated as:
\begin{equation}
p(y_{1} \succ y_{2} \succ y_{3} \mid x) =
\prod_{i=1}^3
\frac{\left(\frac{\pi_\theta(y_{i} \mid x)}{\pi_{\text{ref}}(y_{i} \mid x)}\right)^\beta}
{\sum_{j=i}^3 \left(\frac{\pi_\theta(y_{j} \mid x)}{\pi_{\text{ref}}(y_{j} \mid x)}\right)^\beta}.
\end{equation}

\section{Experiments and results}
\label{Experiments}
In this section, we first introduce our experimental settings, which include the databases, evaluation metrics and implementation details. Secondly, we quantitatively and qualitatively compare the proposed method with fifteen state-of-the-art benchmark MLLMs on the LLSAVisionQA database. Finally, the ablation study is used to validate the effectiveness of each module of our proposed model, as well as the impact of the proposed training strategy.

\subsection{Experimental Settings}

\subsubsection{Database} In order to investigate hallucinations in MLLMs within low-level visual perception and understanding tasks, we propose the HLPU instruction database and LLSAVisionQA database, which are described in detail in Section \ref{Database}. Specifically, the HLPU instruction database is used for training, while the LLSAVisionQA database is used for testing. These databases are employed to validate the effectiveness of the proposed SAFEQA model and the ESA-PO training framework.

\begin{table*}[!t]
    \centering
    \renewcommand\arraystretch{1.05}
    \renewcommand\tabcolsep{2pt}
    \sisetup{table-format=2.3, detect-weight=true, mode=text, table-number-alignment=center} 
    \caption{The quantitative comparison results of the low-level perception ability between different models using $score_{cc}$, $score_{rc}$ and $score_{sa}$, which are tested on the single-image multiple-choice questions in LLSAVisionQA database. Quantitative results are presented according to different types of questions, respectively. 
    ``Base'' in the Table means the untrained model we proposed and we \textbf{bold} the best result and \underline{underlined} the second-best result, the same rule is applied to all tables below.}
    \vspace{-0.5em}
    \label{Question Types1}
    \setstretch{1.1}
    \resizebox{0.95\textwidth}{!}{
    \begin{tabular}{l S S S S S S S S S S S S}
    \toprule[1.5pt]
    Type & \multicolumn{3}{c}{Yes-or-No} & \multicolumn{3}{c}{What} & \multicolumn{3}{c}{How} & \multicolumn{3}{c}{Total}\\ \cmidrule(lr){1-1}
    \cmidrule(lr){2-4} \cmidrule(lr){5-7} \cmidrule(lr){8-10} \cmidrule(lr){11-13}  {Model\textbackslash{}Metirc}
      & {$score_{cc}$↑} & {$score_{rc}$↑} & {$score_{sa}$↑} & {$score_{cc}$↑} & {$score_{rc}$↑} & {$score_{sa}$↑} & {$score_{cc}$↑} & {$score_{rc}$↑} & {$score_{sa}$↑} & {$score_{cc}$↑} & {$score_{rc}$↑} & {$score_{sa}$↑} \\
        \midrule
    LLaVA-Next (8B)\cite{liu2023improved}              & 65.392          & 0.000              & 65.392          & 68.784          & 1.095           & 69.879         & 60.572          & 0.000              & 60.572          & 64.849          & 0.334          & 65.183          \\
    LLaVA-Next (13B)\cite{liu2023improved}              & 65.483          & 0.091          & 65.574          & 60.789          & 1.533           & 62.322         & 59.040           & 0.306           & 59.346          & 61.940           & 0.602          & 62.542          \\
    mPLUG-Owl2 (LLaMA2-7B)\cite{ye2024mplug}             & 65.938          & 0.000             & 65.938          & 54.984          & 4.600             & 59.584         & 56.691          & 2.247           & 58.938          & 59.565          & 2.140           & 61.705          \\
    mPLUG-Owl (Bloomz-7B) \cite{ye2023mplug} & 53.734          & 0.546          & 54.280           & 42.607          & 3.395           & 46.002         & 39.224          & 2.554           & 41.778          & 45.585          & 2.074          & 47.659          \\
    InstructBLIP (Vicuna-7B) \cite{dai2023instructblip}           & 61.566          & 0.000             & 61.566          & 46.440           & 1.095           & 47.535         & 45.455          & 2.962           & 48.417          & 51.672          & 1.304          & 52.976          \\
    InstructBLIP (Flan-T5-XL)\cite{dai2023instructblip}          & 62.204          & 0.273          & 62.477          & 54.217          & 2.738           & 56.955         & 52.809          & 1.124           & 53.933          & 56.689          & 1.304          & 57.993          \\
    InternLM-XComposer2 (7B) \cite{internlmxcomposer2}           & 52.732          & 0.000             & 52.732          & 56.736          & \textbf{10.515} & 67.251         & 45.250           & \textbf{11.951} & 57.201          & 51.505          & \hspace{0.5em}\textbf{7.124} & 58.629          \\
    InternLM-XComposer2d5 (7B) \cite{internlmxcomposer2_5}         & 73.133          & 0.091          & 73.224          & 72.618          & 3.176           & 75.794         & 63.432          & 0.409           & 63.841          & 69.799          & 1.137          & 70.936          \\
    InternLM-XComposer2 (4KHD-7B) \cite{internlmxcomposer2_4khd}      & 68.488          & 0.000             & 68.488          & 74.808          & 0.438           & 75.246         & 65.986          & 0.000              & 65.986          & 69.599          & 0.134          & 69.733          \\
    InfiMM (Zephyr-7B) \cite{InfiMM}                    & 59.927          & 0.000             & 59.927          & 53.779          & 2.081           & 55.860          & 44.433          & 0.919           & 45.352          & 52.977          & 0.936          & 53.913          \\
    Fuyu (Persimmon-8B) \cite{bavishi2023multimodal}                          & 59.199          & 0.091          & 59.290           & 42.388          & 1.533           & 43.921         & 41.369          & 1.430            & 42.799          & 48.227          & 0.970           & 49.197          \\
    Emu2-Chat (LLaMA-33B) \cite{sun2024generative}                        & 68.488          & 0.000             & 68.488          & 61.665          & 4.053           & 65.718         & 44.127          & 6.231           & 50.358          & 58.428          & 3.278          & 61.706          \\
    \hdashline
    GPT-4V (Close-Source) \cite{achiam2023gpt}                            & 68.579          & 0.091          & 68.670           & 70.427          & 1.314           & 71.741         & 60.572          & 1.532           & 62.104          & 66.522          & 0.936          & 67.458          \\
    GPT-4O (Close-Source) \cite{achiam2023gpt}                            & 69.581          & 1.457          & 71.038          & 71.303          & 3.176           & 74.479         & 59.755          & 2.656           & 62.411          & 66.890           & 2.375          & 69.265          \\
    Gemini-1.5-Pro (Close-Source) \cite{google2023geminipro}                   & 58.106          & \hspace{0.5em}\textbf{4.736} & 62.842          & 55.969          & 6.900             & 62.869         & 47.804          & 5.107           & 52.911          & 54.080           & 5.518          & 59.598          \\ \hdashline
    Base                             & \underline{73.497}          & 0.091          & \underline{73.588}          & \underline{76.451}          & 1.972           & \underline{78.423}         & \underline{68.641}          & 0.000              & \underline{68.641}          & \underline{72.809}          & 0.635          & \underline{73.444}          \\
    Proposed                            & \textbf{83.333} & \hspace{0.5em}\underline{3.734}          & \textbf{87.067} & \textbf{83.461} & \hspace{0.5em}\underline{9.529}           & \textbf{92.990} & \textbf{77.017} & \hspace{0.5em}\underline{7.661}           & \textbf{84.678} & \textbf{81.304} & \hspace{0.5em}\underline{6.789} & \textbf{88.093} \\
    \bottomrule[1.5pt]
\end{tabular}
}
\vspace{-1em}
\end{table*}

\subsubsection{Evaluation metrics} We utilize five evaluation metrics to assess hallucinations and response accuracy of MLLMs in low-level visual perception and understanding tasks. One contributing factor to hallucinations is the deficient self-awareness of models, particularly their ability to distinguish between ``known'' and ``unknown'' entities. To evaluate this capability on the LLSAVisionQA database, we first introduce three metrics to measure the self-awareness of MLLMs.

\begin{itemize}
\item $score_{cc}$: It reflects the proportion of questions that the model answers correctly.
\item $score_{rc}$: It represents the proportion of questions that the model appropriately rejects.
\item $score_{sa}$: It is the sum of $score_{cc}$ and $score_{rc}$, representing the overall self-awareness of the model.
\end{itemize}

Before detailing the calculation of these metrics, we introduce some indicators to avoid confusion. For each question $q_i$ in the test set $q$, $c_i$ and $r_i$ represent the indices of the correct answer and the refusal option, respectively. Therefore, $score_{cc}$ and $score_{rc}$ can be defined as:
\begin{equation}
score_{cc} = \frac{100 \cdot \sum_{i=1}^{|q|} \mathbb{I}(p_i = c_i)}{|q|},
\end{equation}
\begin{equation}
score_{rc} = \frac{100 \cdot \sum_{i=1}^{|q|} \mathbb{I}(p_i = r_i) \cdot \mathbb{I}(q_i \text{ is unknown})}{|q|},
\end{equation}
where $p_i$ represents the prediction of the evaluated MLLMs for $q_i$. Since the model may refuse to answer questions that actually knows, when a refusal option is available, we address this issue by removing the refusal option ``I don't know'', forcing the model to select an answer. If the model chooses the correct answer, it demonstrates that the model actually knows the correct response. Consequently, $\mathbb{I}(q_i \text{ is unknown})$ can be defined as follows:
\begin{equation}
\mathbb{I}(q_i \text{ is unknown})=\mathbb{I}(p'_i \neq c_i \mid p_i = r_i),
\end{equation}
where $p'_{i}$ represents the prediction of the model without the refusal option. The self-awareness score($score_{sa}$) is calculated as:
\begin{equation}
score_{sa} = score_{cc} + score_{rc}.
\end{equation}
In order to provide a more comprehensive analysis, we introduce two additional metrics to represent the refusal behavior of MLLMs:
\begin{equation}
\text{Answer Accuracy} = \frac{100 \cdot \sum_{i=1}^{|q|} \mathbb{I}(p_i = c_i)}{\sum_{i=1}^{|q|} \mathbb{I}(p_i \neq r_i)},
\end{equation}
\begin{equation}
\text{SA Rate} = \frac{100 \cdot \sum_{i=1}^{|q|} \mathbb{I}(p_i = r_i)}{|q| - \sum_{i=1}^{|q|} \mathbb{I}(p_i = c_i)},
\end{equation}
where Answer Accuracy represents the proportion of correct predictions among the questions, which the model chooses to answer. The self-awareness (SA) rate represents the rejection probability of the model in answering incorrect questions.

\subsubsection{Experimental details}
We utilize the LLaVA-OneVision as the backbone for both the image feature extraction module and text tokenizer. For the salient region extraction module and quality feature extraction module, we employ the Swin Transformer structure, which are pre-trained on the IQA database \cite{hosu2024uhd}. During the training process, the weights of the text tokenizer are fixed. Initially, we perform supervised finetuning on the proposed SAFEQA model using inputs and positive responses from the HLPU instruction database and subsequently train the model using the proposed ESA-PO framework on HLPU instruction database. All experimental procedures are conducted using eight NVIDIA A100 GPUs to train the model. The learning rate is set to 5$e^{-6}$, the beta value is 0.1 and the number of epochs is 1. We train the model on the training set and test it on the testing set. The testing process is repeated randomly 10 times to mitigate performance variance and the average results are reported for comparison. Our proposed model is implemented with PyTorch.

\begin{table*}[!t]
    \centering
    \renewcommand\arraystretch{1.05}
    \renewcommand\tabcolsep{2pt}
    \sisetup{table-format=2.3, detect-weight=true, mode=text, table-number-alignment=center} 
    \caption{The quantitative comparison results of the low-level perception ability between different models using $score_{cc}$, $score_{rc}$ and $score_{sa}$, which are tested on the single-image multiple-choice questions in LLSAVisionQA database. Quantitative results are presented according to different types of low-level concerns, respectively.}
    \vspace{-0.5em}
    \label{Low-level Concerns1}
    \setstretch{1.1}
    \resizebox{0.95\textwidth}{!}{
    \begin{tabular}{l S S S S S S S S S S S S}
    \toprule[1.5pt]
    Type & \multicolumn{3}{c}{Distortion} & \multicolumn{3}{c}{Other} & \multicolumn{3}{c}{In-context Distortion} & \multicolumn{3}{c}{In-context Other}\\ \cmidrule(lr){1-1}
    \cmidrule(lr){2-4} \cmidrule(lr){5-7} \cmidrule(lr){8-10} \cmidrule(lr){11-13}  {Model\textbackslash{}Metirc}
      & {$score_{cc}$↑} & {$score_{rc}$↑} & {$score_{sa}$↑} & {$score_{cc}$↑} & {$score_{rc}$↑} & {$score_{sa}$↑} & {$score_{cc}$↑} & {$score_{rc}$↑} & {$score_{sa}$↑} & {$score_{cc}$↑} & {$score_{rc}$↑} & {$score_{sa}$↑} \\
        \midrule
    LLaVA-Next (8B)\cite{liu2023improved}              & 55.942          & 0.773           & 56.715          & 69.213          & 0.235          & 69.448          & 63.591          & 0.000             & 63.591          & 77.165          & 0.000             & 77.165          \\
    LLaVA-Next (13B)\cite{liu2023improved}              & 53.720           & 0.386           & 54.106          & 64.395          & 0.823          & 65.218          & 60.906          & 0.671          & 61.577          & 75.787          & 0.591          & 76.378          \\
    mPLUG-Owl2 (LLaMA2-7B)\cite{ye2024mplug}             & 51.884          & 3.961           & 55.845          & 65.452          & 0.705          & 66.157          & 55.034          & 1.678          & 56.712          & 70.669          & 1.378          & 72.047          \\
    mPLUG-Owl (Bloomz-7B) \cite{ye2023mplug} & 37.488          & 2.222           & 39.710           & 48.766          & 2.585          & 51.351          & 46.812          & 1.846          & 48.658          & 55.315          & 1.181          & 56.496          \\
    InstructBLIP (Vicuna-7B) \cite{dai2023instructblip}           & 43.865          & 1.449           & 45.314          & 56.522          & 0.940           & 57.462          & 47.651          & 1.342          & 48.993          & 64.173          & 1.575          & 65.748          \\
    InstructBLIP (Flan-T5-XL)\cite{dai2023instructblip}          & 48.309          & 2.126           & 50.435          & 61.340           & 0.823          & 62.163          & 53.523          & 1.174          & 54.697          & 69.685          & 0.591          & 70.276          \\
    InternLM-XComposer2 (7B) \cite{internlmxcomposer2}           & 42.415          & \textbf{11.498} & 53.913          & 56.287          & \hspace{0.5em}\underline{4.113} & 60.400            & 45.973          & 7.718 & 53.691          & 68.504          & 2.559 & 71.063          \\
    InternLM-XComposer2d5 (7B) \cite{internlmxcomposer2_5}         & 65.217          & 2.512           & 67.729          & 72.385          & 0.470           & 72.855          & 67.282          & 0.336          & 67.618          & 77.756          & 0.394          & 78.150           \\
    InternLM-XComposer2 (4KHD-7B) \cite{internlmxcomposer2_4khd}      & 65.700            & 0.290            & 65.990           & 69.800            & 0.000             & 69.800            & 68.289          & 0.000             & 68.289          & 78.740           & 0.197          & 78.937          \\
    InfiMM (Zephyr-7B) \cite{InfiMM}                    & 42.899          & 2.029           & 44.928          & 58.402          & 0.470           & 58.872          & 49.832          & 0.336          & 50.168          & 68.110           & 0.197          & 68.307          \\
    Fuyu (Persimmon-8B) \cite{bavishi2023multimodal}                          & 41.256          & 1.353           & 42.609          & 52.409          & 0.823          & 53.232          & 45.134          & 0.839          & 45.973          & 59.055          & 0.591          & 59.646          \\
    Emu2-Chat (LLaMA-33B) \cite{sun2024generative}                        & 51.981          & 4.444           & 56.425          & 60.400            & 3.643          & 64.043          & 56.879          & 2.013          & 58.892          & 70.079          & 1.772          & 71.851          \\
    \hdashline
    GPT-4V (Close-Source) \cite{achiam2023gpt}                            & 63.092          & 0.290            & 63.382          & 67.215          & 0.470           & 67.685          & 66.611          & 1.678          & 68.289          & 72.244          & 2.165          & 74.409          \\
    GPT-4O (Close-Source) \cite{achiam2023gpt}                            & 63.188          & 1.353           & 64.541          & 69.565          & 1.763          & 71.328          & 61.745          & 4.698          & 66.443          & 75.984          & 2.756          & 78.74           \\
    Gemini-1.5-Pro (Close-Source) \cite{google2023geminipro}                   & 55.845          & 3.671  & 59.516          & 53.937          & 2.938          & 56.875          & 49.664          & \textbf{10.570} & 60.234          & 55.906          & \hspace{0.5em}\underline{7.677}          & 63.583          \\
    \hdashline
    Base                             & \underline{69.082}          & 1.353           & \underline{70.435}          & \underline{76.498}          & 0.353          & \underline{76.851}          & \underline{68.624}          & 0.336          & \underline{68.960}           & \underline{79.134}          & 0.000             & \underline{79.134}          \\
    Proposed                            & \textbf{79.807} & \hspace{0.5em}\underline{5.604}           & \textbf{85.411} & \textbf{79.906} & \hspace{0.5em}\textbf{5.523} & \textbf{85.429} & \textbf{81.208} & \hspace{0.5em}\underline{9.396}          & \textbf{90.604} & \textbf{86.811} & \hspace{0.5em}\textbf{8.268} & \textbf{95.079} \\
    \bottomrule[1.5pt]
\end{tabular}
}
\vspace{-1em}
\end{table*}

\begin{table*}[!t]
    \centering
    \renewcommand\arraystretch{1}
    \renewcommand\tabcolsep{6pt}
    \sisetup{
      table-format=3.3,       
      detect-weight=true,     
      mode=text,
      detect-inline-weight=math, 
      table-number-alignment=center,
      table-space-text-post = \% 
    }
    \caption{The quantitative comparison results of the low-level perception ability between different models using answer rate and answer accuracy, which are tested on the single-image multiple-choice questions in LLSAVisionQA database. Quantitative results are presented according to different types of questions, respectively. 
    }
    \vspace{-0.5em}
    \label{Question Types2}
    \setstretch{1.2}
    \resizebox{0.8\textwidth}{!}{
    \begin{tabular}{l S S S S S S}
    \toprule[1.5pt]
    Type & \multicolumn{2}{c}{Yes-or-No} & \multicolumn{2}{c}{What} & \multicolumn{2}{c}{How}  \\ \cmidrule(lr){1-1}
    \cmidrule(lr){2-3} \cmidrule(lr){4-5} \cmidrule(lr){6-7}  {Model\textbackslash{}Metirc}
     & {$Answer Acc$↑}    & {$SA Rate$↑}   & {$Answer Acc$↑} & {$SA Rate$↑} & {$Answer Acc$↑} & {$SA Rate$↑} \\ \midrule
    LLaVA-Next (8B)\cite{liu2023improved}                                    & 65.392\%                        & 0.000\%                         & 69.933\%                        & 5.263\%                         & 60.572\%                        & 0.000\%                         \\
    LLaVA-Next (13B)\cite{liu2023improved}                                    & 65.542\%                        & 0.264\%                         & 61.873\%                        & 4.469\%                         & 59.282\%                        & 0.998\%                         \\
    mPLUG-Owl2 (LLaMA2-7B)\cite{ye2024mplug}                                  & 65.938\%                        & 0.000\%                         & 58.035\%                        & 11.679\%                        & 58.980\%                        & 8.962\%                         \\
    mPLUG-Owl (Bloomz-7B) \cite{ye2023mplug}                       & 54.178\%                        & 1.772\%                         & 44.919\%                        & 8.969\%                         & 40.764\%                        & 6.218\%                         \\
    InstructBLIP (Vicuna-7B) \cite{dai2023instructblip}                                 & 61.566\%                        & 0.000\%                         & 47.164\%                        & 2.863\%                         & 47.190\%                        & 6.742\%                         \\
    InstructBLIP (Flan-T5-XL)\cite{dai2023instructblip}                                & 62.431\%                        & 0.964\%                         & 56.059\%                        & 7.177\%                         & 53.575\%                        & 3.030\%                         \\
    InternLM-XComposer2 (7B) \cite{internlmxcomposer2}                                 & 52.732\%                        & 0.000\%                         & 67.448\%                        & \hspace{0.5em}\underline{36.709\%} & 54.691\%                        & \hspace{0.5em}\underline{31.530\%} \\
    InternLM-XComposer2d5 (7B) \cite{internlmxcomposer2_5}                               & 73.200\%                        & 0.339\%                         & 75.341\%                        & 13.200\%                        & 63.758\%                        & 1.397\%                         \\
    InternLM-XComposer2 (4KHD-7B) \cite{internlmxcomposer2_4khd}                           & 68.488\%                        & 0.000\%                         & 75.303\%                        & 2.609\%                         & 65.986\%                        & 0.000\%                         \\
    InfiMM (Zephyr-7B) \cite{InfiMM}                                          & 59.927\%                        & 0.000\%                         & 55.293\%                        & 5.924\%                         & 45.313\%                        & 3.493\%                         \\
    Fuyu (Persimmon-8B) \cite{bavishi2023multimodal}                                                & 59.253\%                        & 0.223\%                         & 43.927\%                        & 6.084\%                         & 41.667\%                        & 1.220\%                         \\
    Emu2-Chat (LLaMA-33B) \cite{sun2024generative}                                              & 68.488\%                        & 0.000\%                         & 66.627\%                        & 19.429\%                        & 48.649\%                        & 16.636\%                        \\ \hdashline
    GPT-4V (Close-Source) \cite{achiam2023gpt}                                                 & 69.019\%                        & 2.029\%                         & 72.492\%                        & 9.630\%                         & 61.900\%                        & 5.440\%                         \\
    GPT-4O (Close-Source) \cite{achiam2023gpt}                                                  & 72.280\%                        & 12.275\%                        & 75.087\%                        & 17.557\%                        & 63.449\%                        & 14.467\%                        \\
    Gemini-1.5-Pro (Close-Source) \cite{google2023geminipro}                                         & 63.800\%                        & \hspace{0.5em}\underline{21.304\%} & 61.198\%                        & 19.403\%                        & 51.599\%                        & 14.090\%                       \\ \hdashline
    Base                                                   & \hspace{0.5em}\underline{73.564\%} & 0.344\%                         & \hspace{0.5em}\underline{78.604\%} & 11.628\%                        & \hspace{0.5em}\underline{68.712\%} & 0.326\%                         \\
    Proposed                                                  & \hspace{0.7em}\textbf{87.476\%}               & \hspace{0.7em}\textbf{28.415\%}               & \hspace{0.7em}\textbf{93.497\%}               & \hspace{0.7em}\textbf{64.901\%}               & \hspace{0.7em}\textbf{86.667\%}               & \hspace{0.7em}\textbf{48.444\%}              \\
    \bottomrule[1.5pt]
\end{tabular}
    }
    \vspace{-1em}
\end{table*}

\begin{table*}[!t]
    \centering
    \renewcommand\arraystretch{1.05}
    \renewcommand\tabcolsep{3pt}
    \sisetup{
      table-format=3.3,       
      detect-weight=true,    
      mode=text,
      detect-inline-weight=math,
      table-number-alignment=center,
      table-space-text-post = \% 
    }
    \caption{The quantitative comparison results of the low-level perception ability between different models using answer rate and answer accuracy, which are tested on the single-image multiple-choice questions in LLSAVisionQA database. Quantitative results are presented according to different types of low-level concerns, respectively.}
    \vspace{-0.5em}
    \label{Low-level Concerns2}
    \setstretch{1.2}
    \resizebox{0.85\textwidth}{!}{
    \begin{tabular}{l S S S S S S S S}
    \toprule[1.5pt]
    Type  & \multicolumn{2}{c}{Distortion} & \multicolumn{2}{c}{Other} & \multicolumn{2}{c}{In-context Distortion} & \multicolumn{2}{c}{In-context Other} \\  \cmidrule(lr){1-1}  
    \cmidrule(lr){2-3} \cmidrule(lr){4-5} \cmidrule(lr){6-7} \cmidrule(lr){8-9}  {Model\textbackslash{}Metirc}
    & {$Answer Acc$↑}    & {$SA Rate$↑}   & {$Answer Acc$↑} & {$SA Rate$↑} & {$Answer Acc$↑} & {$SA Rate$↑} & {$Answer Acc$↑} & {$SA Rate$↑} \\ \midrule
    LLaVA-Next (8B)\cite{liu2023improved}                                    & 56.488\%                        & 2.193\%                         & 69.458\%                        & 1.145\%                         & 63.805\%                        & 0.922\%                         & 77.165\%                        & 0.000\%                         \\
    LLaVA-Next (13B)\cite{liu2023improved}                                    & 53.928\%                        & 0.835\%                         & 65.161\%                        & 3.300\%                         & 61.318\%                        & 1.717\%                         & 76.238\%                        & 2.439\%                         \\
    mPLUG-Owl2 (LLaMA2-7B)\cite{ye2024mplug}                                  & 54.518\%                        & 10.040\%                        & 66.389\%                        & 4.082\%                         & 56.164\%                        & 4.478\%                         & 72.379\%                        & 8.054\%                         \\
    mPLUG-Owl (Bloomz-7B) \cite{ye2023mplug}                       & 38.761\%                        & 5.255\%                         & 50.610\%                        & 7.110\%                         & 48.021\%                        & 4.732\%                         & 56.768\%                        & 5.727\%                         \\
    InstructBLIP (Vicuna-7B) \cite{dai2023instructblip}                                 & 44.553\%                        & 2.754\%                         & 57.467\%                        & 3.784\%                         & 48.464\%                        & 3.205\%                         & 65.462\%                        & 5.495\%                         \\
    InstructBLIP (Flan-T5-XL)\cite{dai2023instructblip}                                & 49.505\%                        & 4.673\%                         & 61.922\%                        & 2.432\%                         & 54.530\%                        & 3.971\%                         & 70.238\%                        & 2.597\%                         \\
    InternLM-XComposer2 (7B) \cite{internlmxcomposer2}                                 & 50.518\%                        & \hspace{0.5em}\underline{27.852\%} & 60.252\%                        & 15.054\%                        & 52.290\%                        & 22.360\%                        & 71.311\%                        & 12.500\%                        \\
    InternLM-XComposer2d5 (7B) \cite{internlmxcomposer2_5}                               & 67.097\%                        & 8.056\%                         & 72.727\%                        & 1.702\%                         & 67.622\%                        & 1.538\%                         & 78.218\%                        & 2.655\%                         \\
    InternLM-XComposer2 (4KHD-7B) \cite{internlmxcomposer2_4khd}                           & 66.019\%                        & 1.408\%                         & 69.800\%                        & 0.000\%                         & 68.289\%                        & 0.000\%                         & 78.895\%                        & 0.926\%                         \\
    InfiMM (Zephyr-7B) \cite{InfiMM}                                          & 44.004\%                        & 4.399\%                         & 59.096\%                        & 2.825\%                         & 50.254\%                        & 1.672\%                         & 68.515\%                        & 1.852\%                         \\
    Fuyu (Persimmon-8B) \cite{bavishi2023multimodal}                                                & 42.152\%                        & 3.618\%                         & 52.906\%                        & 1.975\%                         & 45.593\%                        & 1.835\%                         & 59.524\%                        & 1.923\%                         \\
    Emu2-Chat (LLaMA-33B) \cite{sun2024generative}                                              & 55.751\%                        & 14.085\%                        & 64.250\%                        & \hspace{0.5em}\underline{15.134\%} & 58.957\%                        & 8.171\%                         & 72.505\%                        & 11.184\%                        \\ \hdashline
    GPT-4V (Close-Source) \cite{achiam2023gpt}                                                 & 63.460\%                        & 1.571\%                         & 67.612\%                        & 1.792\%                         & \hspace{0.5em}\underline{69.527\%} & 12.563\%                        & 74.898\%                        & 12.766\%                        \\
    GPT-4O (Close-Source) \cite{achiam2023gpt}                                                  & 65.010\%                        & 7.612\%                         & 72.906\%                        & 15.058\%                        & 68.022\%                        & 24.123\%                        & \hspace{0.5em}\underline{79.261\%} & 17.213\%                        \\
    Gemini-1.5-Pro (Close-Source) \cite{google2023geminipro}                                         & 58.680\%                        & 10.941\%                        & 56.877\%                        & 11.224\%                        & 59.082\%                        & \hspace{0.5em}\underline{31.667\%} & 63.252\%                        & \hspace{0.5em}\underline{26.339\%} \\ \hdashline
    Base                                                   & \hspace{0.5em}\underline{70.722\%} & 7.500\%                         & \hspace{0.5em}\underline{76.769\%} & 1.500\%                         & 68.624\%                        & 0.000\%                         & 79.134\%                        & 0.000\%                         \\
    Proposed                                                  & \hspace{0.7em}\textbf{86.674\%}               & \hspace{0.7em}\textbf{39.234\%}               & \hspace{0.7em}\textbf{85.106\%}               & \hspace{0.7em}\textbf{30.409\%}               & \hspace{0.7em}\textbf{93.617\%}               & \hspace{0.7em}\textbf{70.536\%}               & \hspace{0.7em}\textbf{95.455\%}               & \hspace{0.7em}\textbf{68.657\%} \\
    \bottomrule[1.5pt]
\end{tabular}
    }
    \vspace{-1em}
\end{table*}

\subsection{Comparison and Analysis of Self-awareness Ability}
We evaluate the hallucination of our proposed method in low-level visual perception and understanding tasks on the LLSAVisionQA database. We compare the performance of the proposed SAFEQA model with fifteen popular MLLMs including twelve open-source models: LLaVA-Next (8B)\cite{liu2023improved}, LLaVA-Next (13B)\cite{liu2023improved}, mPLUG-Owl2 (LLaMA2-7B)\cite{ye2024mplug}, mPLUG-Owl (Bloomz-7B) \cite{ye2023mplug}, InstructBLIP (Vicuna-7B) \cite{dai2023instructblip}, InstructBLIP (Flan-T5-XL)\cite{dai2023instructblip}, InternLM-XComposer2 (7B) \cite{internlmxcomposer2}, InternLM-XComposer2d5 (7B) \cite{internlmxcomposer2_5}, InternLM-XComposer2 (4KHD-7B) \cite{internlmxcomposer2_4khd}, InfiMM (Zephyr-7B) \cite{InfiMM}, Fuyu (Persimmon-8B) \cite{bavishi2023multimodal}, Emu2-Chat (LLaMA-33B) \cite{sun2024generative} and three close-source models: GPT-4V \cite{achiam2023gpt}, GPT-4O \cite{achiam2023gpt}. Gemini-1.5-Pro \cite{google2023geminipro}.

Table \ref{Question Types1} presents the metrics for different problem types associated with the low-level visual perception and understanding tasks of both the baseline models and our proposed model, including the $score_{cc}$, $score_{rc}$ and $score_{sa}$. First, our proposed model demonstrates the most significant improvement in the ``Yes-or-No’’ questions with $score_{cc}$, compared to the baseline MLLMs, with the $score_{cc}$ surpassing the base model by 9.836\%. In terms of ``How" questions, the $score_{rc}$ and $score_{sa}$ of our proposed model achieves the most significant improvement over the base model, with an increase of 7.661\% and 16.037\%. Additionally, compared to baseline MLLMs, the $score_{sa}$ of our proposed model consistently outperforms that of the suboptimal model across all problem types. Moreover, all metrics of our proposed model surpass those of GPT-4O, with an overall $score_{sa}$ that exceeds GPT-4O by 18.828\%. Finally, among the currently popular MLLMs, InternLM-XComposer2d5 and InternLM-XComposer2 (4KHD) achieve superior results in the $score_{cc}$, while InternLM-XComposer2 outperforms in the $score_{rc}$. However, a comprehensive analysis of the data reveals that the current popular MLLMs struggle to achieve a balanced performance across both the $score_{cc}$ and $score_{rc}$. This suggests they lack a clear understanding of the knowledge boundaries in low-level vision tasks, which contributes to their lower performance on the $score_{sa}$.

\begin{table*}[!t]
    \centering
    \renewcommand\arraystretch{1}
    \renewcommand\tabcolsep{2pt}
    \sisetup{table-format=2.3, detect-weight=true, mode=text, table-number-alignment=center} 
    \caption{Ablation studies on extraction modules of our proposed model, SAFEQA.}
    \vspace{-0.5em}
    \label{extract module}
    \setstretch{1.1}
    \resizebox{0.95\textwidth}{!}{
    \begin{tabular}{l S S S S S S S S S S S S S S}
    \toprule[1.5pt]
    Type & \multicolumn{2}{c}{Yes-or-no} & \multicolumn{2}{c}{What} & \multicolumn{2}{c}{How} & \multicolumn{2}{c}{Distortion} & \multicolumn{2}{c}{Other} & \multicolumn{2}{c}{In-context Distortion} & \multicolumn{2}{c}{In-context Other}\\ \cmidrule(lr){1-1} 
    \cmidrule(lr){2-3} \cmidrule(lr){4-5} \cmidrule(lr){6-7} \cmidrule(lr){8-9} \cmidrule(lr){10-11} \cmidrule(lr){12-13} \cmidrule(lr){14-15} {Model\textbackslash{}Metirc}
        & {$score_{cc}$↑} & {$score_{rc}$↑} & {$score_{cc}$↑} & {$score_{rc}$↑} & {$score_{cc}$↑} & {$score_{rc}$↑} & {$score_{cc}$↑} & {$score_{rc}$↑} & {$score_{cc}$↑} & {$score_{rc}$↑} & {$score_{cc}$↑} & {$score_{rc}$↑} & {$score_{cc}$↑} & {$score_{rc}$↑} \\
        \midrule
Base                     & 73.497                        & 0.091                        & 76.451                        & 1.972                        & 68.641                        & 0.000                        & 69.082                        & 1.353                        & 76.498                        & 0.353                        & 68.624                        & 0.336                        & 79.134                        & 0.000                        \\
\emph{w/o} SFT                  & 76.138                        & 1.913                        & 76.561                        & 5.148                        & 69.867                        & 3.984                        & 72.271                        & 2.802                        & 75.558                        & 2.585                        & 68.960                        & 5.201                        & 82.087                        & 4.921                        \\
\emph{w/o} quality              & 80.055                        & 3.097                        & 78.313                        & 6.900                        & \underline{74.668} & 5.924                        & 74.879                        & 3.478                        & \underline{78.026} & 3.878                        & \underline{76.510} & 8.054                        & 84.646                        & \hspace{0.5em}\underline{7.480} \\
\emph{w/o} saliency             & \underline{80.510} & \hspace{0.5em}\underline{3.461} & \underline{78.642} & \hspace{0.5em}\underline{7.777} & 73.442                        & \hspace{0.5em}\underline{6.333} & \underline{75.266} & \hspace{0.5em}\underline{4.638} & 76.381                        & \hspace{0.5em}\underline{4.113} & 76.007                        & \hspace{0.5em}\underline{9.228} & \underline{86.417} & 6.496                        \\
Proposed                 & \textbf{83.333}               & \hspace{0.5em}\textbf{3.734}               & \textbf{83.461}               & \hspace{0.5em}\textbf{9.529}               & \textbf{77.017}               & \hspace{0.5em}\textbf{7.661}               & \textbf{79.807}               & \hspace{0.5em}\textbf{5.604}               & \textbf{79.906}               & \hspace{0.5em}\textbf{5.523}               & \textbf{81.208}               & \hspace{0.5em}\textbf{9.396}               & \textbf{86.811}               & \hspace{0.5em}\textbf{8.268} 
\\
    \bottomrule[1.5pt]      
\end{tabular}
}
\vspace{-1em}
\end{table*}

Table \ref{Low-level Concerns1} presents the metrics for various distortion types associated with the low-level visual perception and understanding tasks of the same baseline models and our proposed model, including the $score_{cc}$, $score_{rc}$ and $score_{sa}$. First of all, it can be observed that when it comes to distortion and in-context distortion, the $score_{cc}$ of the proposed model is significantly improved compared with that of the base model, which increases by 10.725\% and 12.584\%, respectively. Additionally, the average $score_{cc}$ and $score_{rc}$ across the four distortion types is approximately 14.414\% and 4.414\% higher than GPT-4O. A comparative analysis of both tables reveals that the model exhibits low self-awareness capabilities in relatively simple problems, while demonstrating high self-awareness capabilities in more complex problems. 

\subsection{Comparison and Analysis of Rejection Behavior}
To facilitate a more detailed analysis, we introduce two additional metrics to represent the rejection behavior of MLLMs: answer accuracy and self-awareness rate. Experimental results are shown in Table \ref{Question Types2} and Table \ref{Low-level Concerns2}. In terms of different problem types, our proposed model shows the most significant improvement in answer accuracy for ``How’’ questions and self-awareness rate for ``What" questions when compared to the baseline model. Regarding different distortion types, our model shows the greatest improvement in answer accuracy and self-awareness rate under in-context distortion, with an improvement of 25.595\% in answer accuracy compared to GPT-4O. Across all subsets, both the answer accuracy index and self-awareness rate index of our proposed method outperform those of the current popular MLLMs.

\begin{figure*}[!t]
\centering
\includegraphics[width=6.5in]{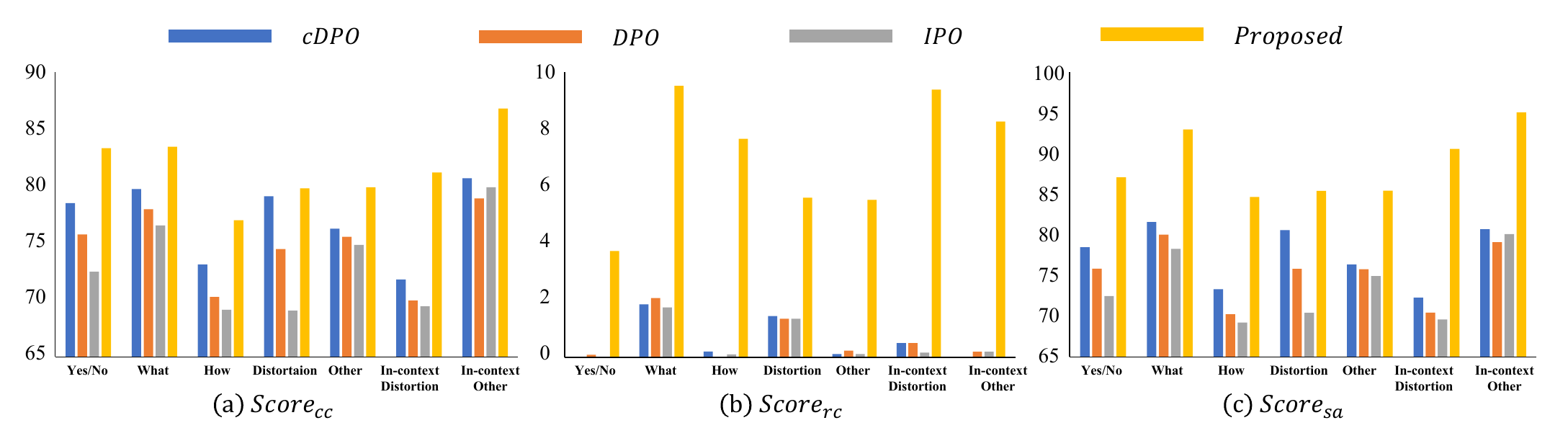}
\vspace{-1em}
\caption{Comparison of the low-level perception ability between the SAFEQA model trained on different preference optimization algorithms.}
\label{PO Method}
\vspace{-1em}
\end{figure*}

\begin{table*}[!t]
    \centering
    \renewcommand\arraystretch{1}
    \renewcommand\tabcolsep{2pt}
    \sisetup{table-format=2.3, detect-weight=true, mode=text, table-number-alignment=center} 
    \caption{Comparison of the low-level perception ability between different database partitions.}
    \vspace{-0.5em}
    \label{combinations of data}
    \setstretch{1.1}
    \resizebox{0.95\textwidth}{!}{
    \begin{tabular}{l S S S S S S S S S S S S S S}
    \toprule[1.5pt]
    Type & \multicolumn{2}{c}{Yes-or-no} & \multicolumn{2}{c}{What} & \multicolumn{2}{c}{How} & \multicolumn{2}{c}{Distortion} & \multicolumn{2}{c}{Other} & \multicolumn{2}{c}{In-context Distortion} & \multicolumn{2}{c}{In-context Other}\\ \cmidrule(lr){1-1}
    \cmidrule(lr){2-3} \cmidrule(lr){4-5} \cmidrule(lr){6-7} \cmidrule(lr){8-9} \cmidrule(lr){10-11} \cmidrule(lr){12-13} \cmidrule(lr){14-15} {Model\textbackslash{}Metirc}
        & {$score_{cc}$↑} & {$score_{rc}$↑} & {$score_{cc}$↑} & {$score_{rc}$↑} & {$score_{cc}$↑} & {$score_{rc}$↑} & {$score_{cc}$↑} & {$score_{rc}$↑} & {$score_{cc}$↑} & {$score_{rc}$↑} & {$score_{cc}$↑} & {$score_{rc}$↑} & {$score_{cc}$↑} & {$score_{rc}$↑} \\
        \midrule
    Base                     & 73.497                        & 0.091                        & 76.451                        & 1.972                        & 68.641                        & 0.000                           & 69.082                        & 1.353                        & 76.498                        & 0.353                        & 68.624                        & 0.336                        & 79.134                        & 0.000                           \\
    Conversation       & \underline{78.689} & \hspace{0.5em}\underline{3.188} & 80.175                        & 7.010                         & \underline{75.689} & \hspace{0.5em}\underline{5.209} & \underline{76.135} & \hspace{0.5em}\underline{5.217} & \underline{78.026} & 2.350                         & \underline{76.007} & \hspace{0.5em}\underline{8.054} & \underline{85.039} & \hspace{0.5em}\underline{5.512} \\
    Yes-or-No          & 76.958                        & 2.823                        & 76.232                        & 5.367                        & 68.641                        & 4.086                        & 69.662                        & 4.058                        & 75.558                        & 2.115                        & 70.805                        & 6.711                        & 84.055                        & 3.937                        \\
    What/How          & 75.774                        & 2.277                        & \underline{81.709} & \hspace{0.5em}\underline{7.448} & 72.217                        & 4.290                        & 73.430                         & 4.348                        & 76.146                        & \hspace{0.5em}\underline{2.938} & 75.336                        & 6.879                        & 84.252                        & 4.724                        \\
    Proposed                 & \textbf{83.333}               & \hspace{0.5em}\textbf{3.734}               & \textbf{83.461}               & \hspace{0.5em}\textbf{9.529}               & \textbf{77.017}               & \hspace{0.5em}\textbf{7.661}               & \textbf{79.807}               & \hspace{0.5em}\textbf{5.604}               & \textbf{79.906}               & \hspace{0.5em}\textbf{5.523}               & \textbf{81.208}               & \hspace{0.5em}\textbf{9.396}               & \textbf{86.811}               & \hspace{0.5em}\textbf{8.268}
   \\ 
    \bottomrule[1.5pt]      
\end{tabular}
}
\vspace{-1em}
\end{table*}

\subsection{Ablation Study}

In this section, we conduct ablation studies and provide corresponding analyses, including the impact of various modules in the proposed model, the preference optimization method and the database partitions.

\subsubsection{Contribution of Low-level Visual Feature Extraction Module}
To verify the rationality and effectiveness of the salient region feature extraction module and the quality feature extraction module in the proposed SAFEQA model, we conduct ablation experiments from the following two aspects: including \emph{w/o} saliency, which means removing the salient region feature extraction module and \emph{w/o} quality, which means removing the quality feature extraction module. Table \ref{extract module} presents the results of this ablation study, indicating that these two modules are indispensable, highlighting the positive impact of these two modules on the performance across different tasks and evaluation indicators. Furthermore, we also include \emph{w/o} supervised finetuning, which means that before employing the ESA-PO training framework, the proposed model does not utilize the images and positive responses from the HLPU instruction database for supervised finetuning. The experimental results in Table \ref{extract module} demonstrate that supervised finetuning before preference optimization can improve the overall performance of the model.

\subsubsection{Effects of Optimization Algorithms}
We empirically compare the effectiveness of several optimization algorithms: (1) DPO \cite{rafailov2024direct}, which directly finetunes the model on offline preference databases without explicitly constructing a reward function. (2) cDPO \cite{mitchell2023note}, a variant of the DPO loss that accounts for potential label noise in preference data. (3) IPO \cite{azar2024general}, which introduces an improved loss function to mitigate overfitting in DPO by employing the average log likelihood ratio and controlling the gap between selection and rejection completions via beta parameters. For all algorithms, we set the learning rate to 5$e^{-6}$ and use the corresponding hyperparameters recommended in the respective papers. We evaluate these algorithms on LLSAVisionQA database and select some metrics for presentation. The visualization results shown in Fig. \ref{PO Method}, clearly demonstrate that our proposed ESA-PO algorithm significantly enhances the self-awareness of model.

\subsubsection{Comparison Between Different Database Partitions}
In the low-level visual perception and understanding tasks, we combine subsets of different question and answer types together and train them jointly under our unified preference optimization method. To validate the effectiveness of each subset, we implement separate preference optimization for each subset. The results are shown in Table \ref{combinations of data}, which demonstrate that each subset enhances model performance and effectively mitigates the hallucinations. Notably, the subsets corresponding to different problem types exhibit significant improvements on their respective test sets and other problem types also benefit from the optimization compared to the base model. The conversations subset demonstrates the most substantial average improvement compared to the base model, attributed to its rich informational content.

\section{Conclusion}
\label{Conclusion}
Low-level visual perception and understanding is an important task in the fields of computer vision and image processing. In this work, we conduct an in-depth exploration of the phenomenon of hallucinations in low-level visual perception and understanding tasks. Specifically, we consider that a significant factor contributing to the hallucinations of models is the absence of self-awareness. To address this, we construct the first database focusing on hallucination phenomena in low-level visual perception and understanding tasks termed HLPU instruction database, where each image and its corresponding question have multiple different answers reflecting different preferences. Our constructed HLPU instruction database is divided into four subsets, encompassing various types of distortions and contains a total of 200K images with corresponding preference data. Through qualitative and quantitative analysis, we conclude that existing MLLMs lack sufficient self-awareness capabilities and demonstrate varying degrees of hallucination across different types of questions and distortions. A novel model, termed SAFEQA, is then proposed to better handle the low-level visual perception and understanding tasks, which integrates image features, salient region features and quality features to enhance the ability of the model in perceiving and understanding low-level features. Furthermore, to improve the self-awareness of models, we propose the ESA-PO, which enables the model to learn human preferences more effectively. Experimental results on the LLSAVisionQA database validate that our proposed model and training strategy outperform close-source models in both self-awareness ability and answer accuracy, demonstrating the superiority of our method.

\bibliographystyle{IEEEtran}
\bibliography{ref}

\vspace{-5em}
\begin{IEEEbiography}[{\includegraphics[width=1in,height=1.25in,clip,keepaspectratio]{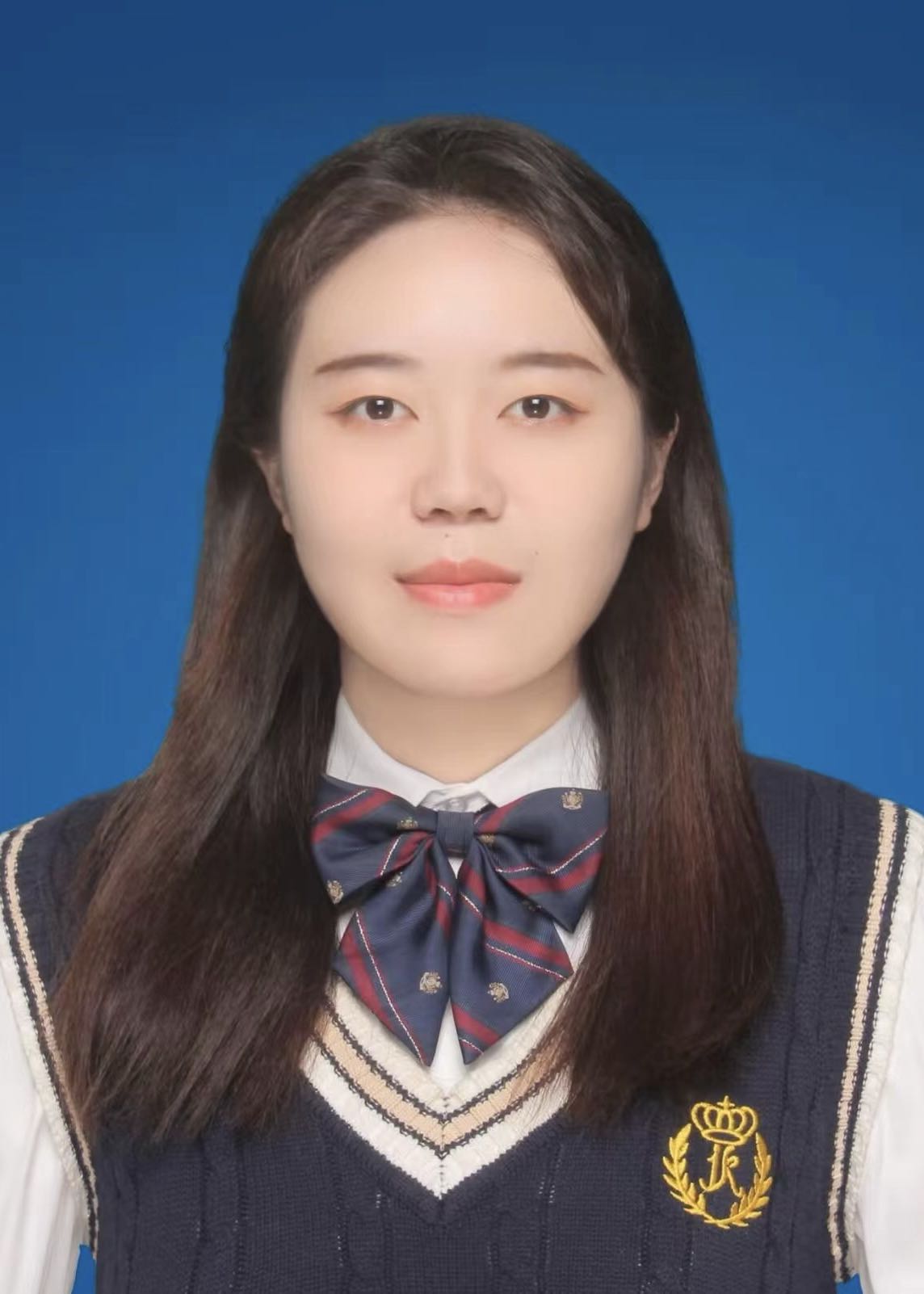}}]{Yinan Sun}
received the B.E. degree from Wuhan University, Wuhan, China, in 2022. She is currently pursuing her Ph.D. degree at the School of Electronic Information and Electrical Engineering at Shanghai Jiao Tong University, Shanghai, China. Her research interests include multimodal visual attention modeling and multimedia signal processing.
\end{IEEEbiography}

\vspace{-5em}
\begin{IEEEbiography}[{\includegraphics[width=1in,height=1.25in,clip,keepaspectratio]{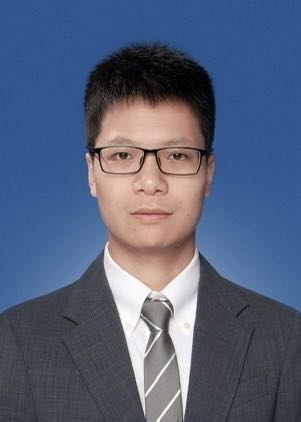}}]{Xiongkuo Min}
(Member, IEEE) received the B.E. degree from Wuhan University, Wuhan, China, in 2013, and the Ph.D. degree from Shanghai Jiao Tong University, Shanghai, China, in 2018, where he is currently a tenure-track Associate Professor with the Institute of Image Communication and Network Engineering. His research interests include image/video/audio quality assessment, quality of experience, visual attention modeling, extended reality, and multimodal signal processing. 
\end{IEEEbiography}

\vspace{-5em}
\begin{IEEEbiography}[{\includegraphics[width=1in,height=1.25in,clip,keepaspectratio]{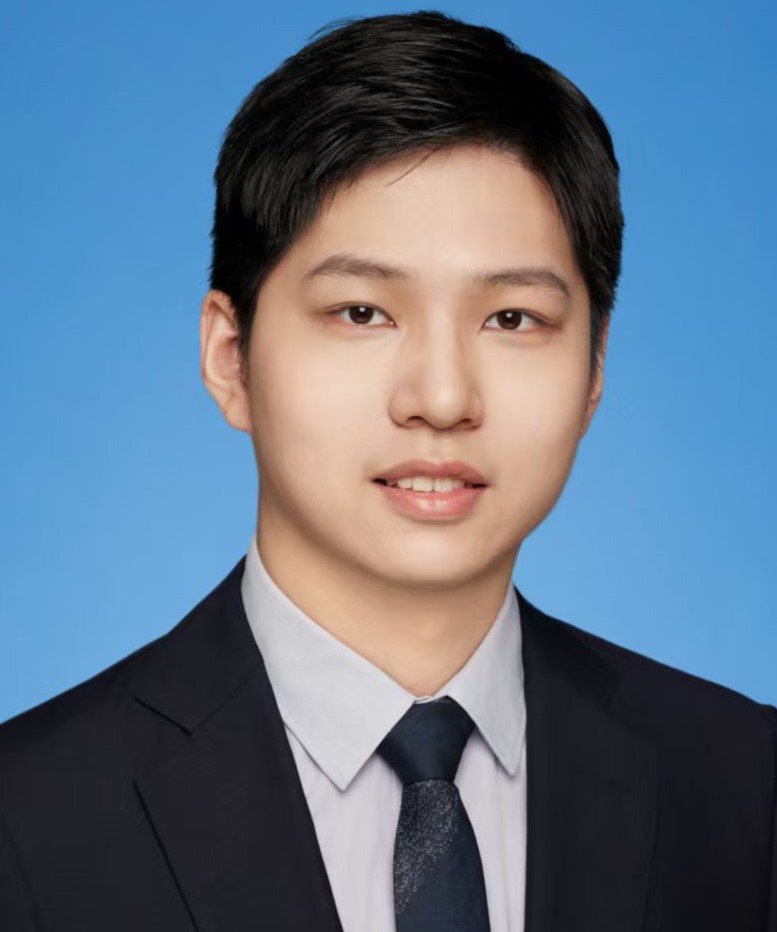}}]{Zicheng Zhang}
received his B.E. degree from Shanghai Jiaotong University, Shanghai, China, in 2020 and he is currently pursuing PhD degree at Shanghai Jiao Tong University. His research interests include quality assessment, low-level vision, and large multi-modal models.
\end{IEEEbiography}

\vspace{-5em}
\begin{IEEEbiography}[{\includegraphics[width=1in,height=1.25in,clip,keepaspectratio]{
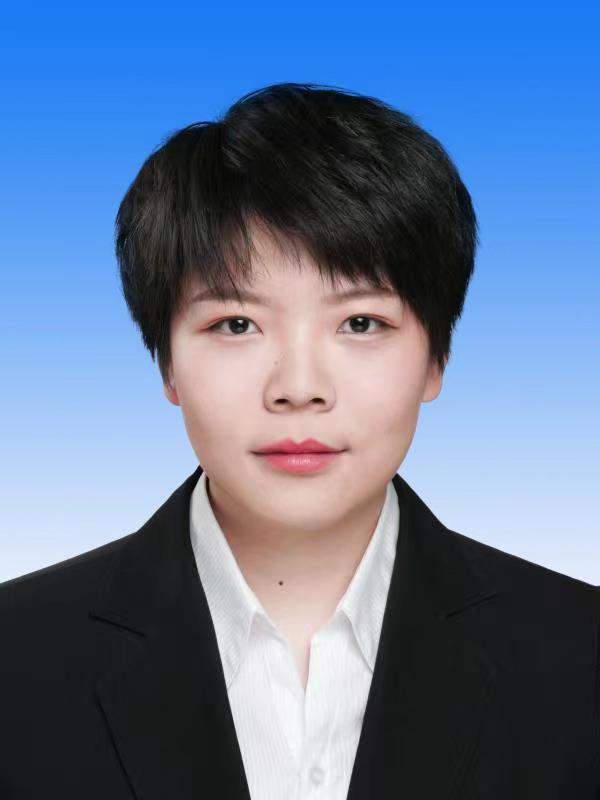}}]{Yixuan Gao}
received the B.E. degree from the Harbin Institute of Technology, Weihai, China, in 2020. She is currently working toward a Ph.D. degree with the Institute of Image Communication and Network Engineering, Shanghai Jiao Tong University, Shanghai, China. Her current research interest is in image quality assessment.
\end{IEEEbiography}

\vspace{-5em}
\begin{IEEEbiography}[{\includegraphics[width=1in,height=1.25in,clip,keepaspectratio]{
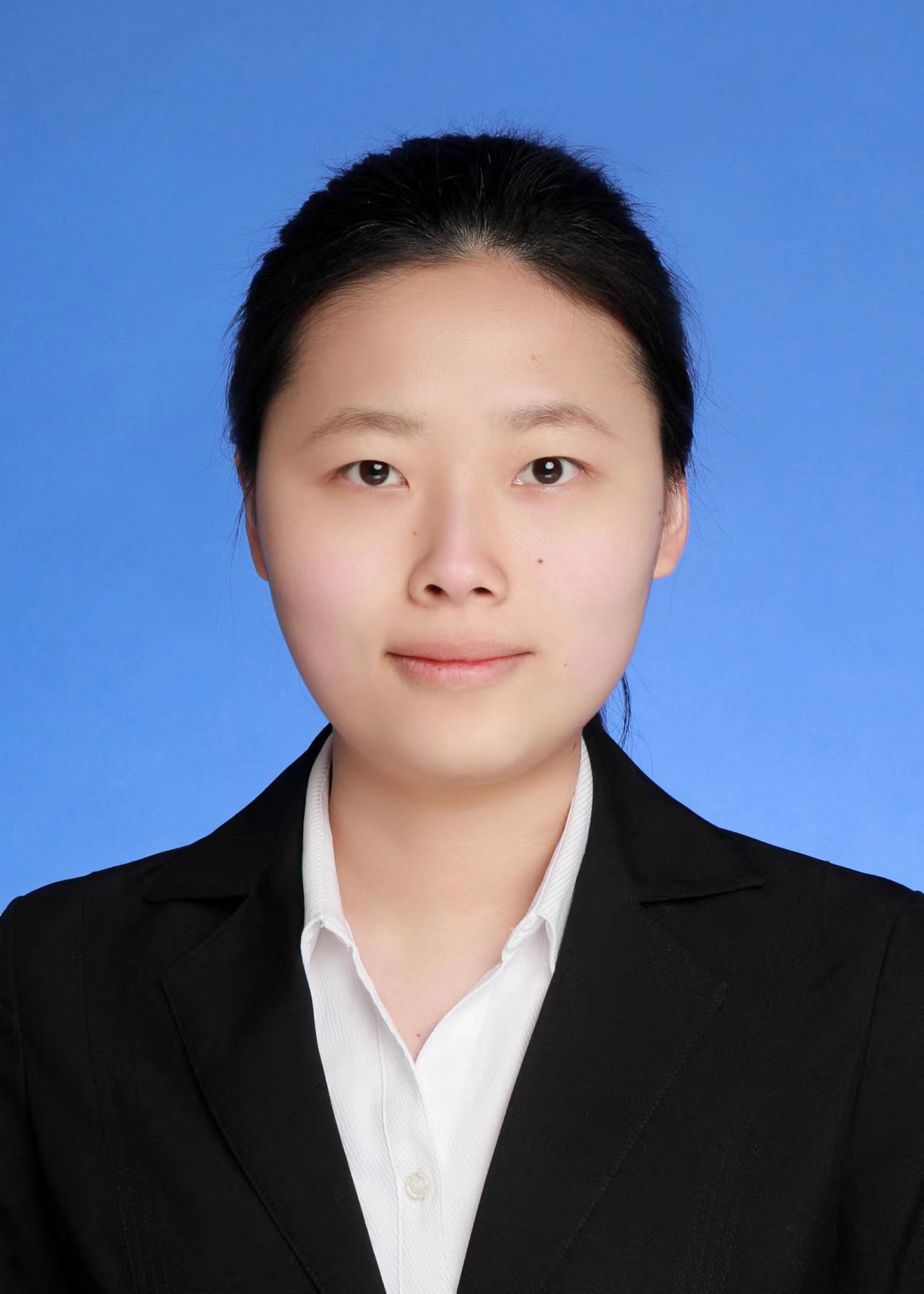}}]{Yuqin Cao}
received the B.E. degree from the Shanghai Jiao Tong University, Shanghai, China, in 2021. She is currently working toward a Ph.D. degree with the Institute of Image Communication and Network Engineering, Shanghai Jiao Tong University, Shanghai, China. Her current research interest is in audio-visual quality assessment.
\end{IEEEbiography}

\vspace{-5em}
\begin{IEEEbiography}[{\includegraphics[width=1in,height=1.25in,clip,keepaspectratio]{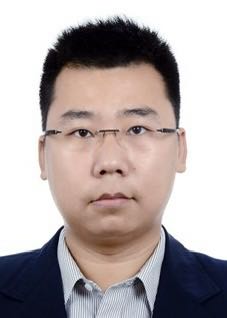}}]{Guangtao Zhai}
(Fellow, IEEE) received the B.E. and M.E. degrees from Shandong University, in 2001 and 2004, respectively, and the Ph.D. degree from Shanghai Jiao Tong University, Shanghai, China, in 2009. He is currently a Research Professor with the Institute of Image Communication and Information Processing, Shanghai Jiao Tong University. His research interests include multimedia signal processing and perceptual signal processing. He received the Award of National Excellent Ph.D. Thesis from the Ministry of Education of China in 2012.
\end{IEEEbiography}

\end{document}